# Relaxed Survey Propagation for
# The Weighted Maximum Satisfiability Problem

**Hai Leong Chieu**                                              CHAILEON@DSO.ORG.SG
*DSO National Laboratories,*
*20 Science Park Drive, Singapore 118230*

**Wee Sun Lee**                                                 LEEWS@COMP.NUS.EDU.SG
*Department of Computer Science, School of Computing,*
*National University of Singapore, Singapore 117590*

## Abstract

The survey propagation (SP) algorithm has been shown to work well on large instances of the random 3-SAT problem near its phase transition. It was shown that SP estimates marginals over *covers* that represent clusters of solutions. The SP-$y$ algorithm generalizes SP to work on the maximum satisfiability (Max-SAT) problem, but the *cover* interpretation of SP does not generalize to SP-$y$. In this paper, we formulate the relaxed survey propagation (RSP) algorithm, which extends the SP algorithm to apply to the weighted Max-SAT problem. We show that RSP has an interpretation of estimating marginals over *covers* violating a set of clauses with minimal weight. This naturally generalizes the *cover* interpretation of SP. Empirically, we show that RSP outperforms SP-$y$ and other state-of-the-art Max-SAT solvers on random Max-SAT instances. RSP also outperforms state-of-the-art weighted Max-SAT solvers on random weighted Max-SAT instances.

## 1. Introduction

The 3-SAT problem is the archetypical NP-complete problem, and the difficulty of solving random 3-SAT instances has been shown to be related to the clause to variable ratio, $\alpha = M/N$, where $M$ is the number of clauses and $N$ the number of variables. A phase transition occurs at the critical value of $\alpha_c \approx 4.267$: random 3-SAT instances with $\alpha < \alpha_c$ are generally satisfiable, while instances with $\alpha > \alpha_c$ are not. Instances close to the phase transition are generally hard to solve using local search algorithms (Mezard & Zecchina, 2002; Braunstein, Mezard, & Zecchina, 2005).

The survey propagation (SP) algorithm was invented in the statistical physics community using approaches used for analyzing phase transitions in spin glasses (Mezard & Zecchina, 2002). The SP algorithm has surprised computer scientists by its ability to solve efficiently extremely large and difficult Boolean satisfiability (SAT) instances in the phase transition region. The algorithm has also been extended to the SP-$y$ algorithm to handle the maximum satisfiability (Max-SAT) problem (Battaglia, Kolar, & Zecchina, 2004).

Progress has been made in understanding why the SP algorithm works well. Braunstein and Zecchina (2004) first showed that SP can be viewed as the belief propagation (BP) algorithm (Pearl, 1988) on a related factor graph where only clusters of solutions represented by covers have non-zero probability. It is not known whether a similar interpretation can be given to the SP-$y$ algorithm. In this paper, we extend the SP algorithm to handle weighted

Max-SAT instances in a way that preserves the cover interpretation, and we call this new algorithm the Relaxed Survey Propagation (RSP) algorithm. Empirically, we show that RSP outperforms SP-$y$ and other state-of-the-art solvers on random Max-SAT instances. It also outperforms state-of-the-art solvers on a few benchmark Max-SAT instances. On random weighted Max-SAT instances, it outperforms state-of-the-art weighted Max-SAT solvers.

The rest of this paper is organized as follows. In Section 2, we describe the background literature and mathematical notations necessary for understanding this paper. This includes a brief review of the definition of joint probability distributions over factor graphs, an introduction to the SAT, Max-SAT and the weighted Max-SAT problem, and how they can be formulated as inference problems over a probability distribution on a factor graph. In Section 3, we give a review of the BP algorithm (Pearl, 1988), which plays a central role in this paper. In Section 4, we give a description of the SP (Braunstein et al., 2005) and the SP-$y$ (Battaglia et al., 2004) algorithm, explaining them as warning propagation algorithms. In Section 5, we define a joint distribution over an extended factor graph given a weighted Max-SAT instance. This factor graph generalizes the factor graph defined by Maneva, Mossel and Wainwright (2004) and by Chieu and Lee (2008). We show that, for solving SAT instances, running the BP algorithm on this factor graph is equivalent to running the SP algorithm derived by Braunstein, Mezard and Zecchina (2005). For the weighted Max-SAT problem, this gives rise to a new algorithm that we call the Relaxed Survey Propagation (RSP) algorithm. In Section 7, we show empirically that RSP outperforms other algorithms for solving hard Max-SAT and weighted Max-SAT instances.

## 2. Background

While SP was first derived from principles in statistical physics, it can be understood as a BP algorithm, estimating marginals for a joint distribution defined over a factor graph. In this section, we will provide background material on joint distributions defined over factor graphs. We will then define the Boolean satisfiability (SAT) problem, the maximum satisfiability (Max-SAT) problem, and the weighted maximum satisfiability (weighted Max-SAT) problem, and show that these problems can be solved by solving an inference problem over joint distributions defined on factor graphs. A review of the definition and derivation of the BP algorithm will then follow in the next section, before we describe the SP algorithm in Section 4.

### 2.1 Notations

First, we will define notations and concepts that are relevant to the inference problems over factor graphs. Factor graphs provide a framework for reasoning and manipulating the joint distribution over a set of variables. In general, variables could be continuous in nature, but in this paper, we limit ourselves to discrete random variables.

In this paper, we denote random variables using large Roman letters, e.g., $X, Y$. The random variables are always discrete in this paper, taking values in a finite domain. Usually, we are interested in vectors of random variables, for which we will write the letters in bold face, e.g., $\mathbf{X}, \mathbf{Y}$. We will often index random variables by the letters $i, j, k...$, and write, for example, $\mathbf{X} = \{X_i\}_{i \in V}$, where $V$ is a finite set. For a subset $W \subseteq V$, we will denote $\mathbf{X}_W = \{X_i\}_{i \in W}$. We call an assignment of values to the variables in $\mathbf{X}$ a *configuration*, and will denote it in small bold letters, e.g. $\mathbf{x}$. We will often write $x$ to represent an event $X = x$ and, for a probability distribution $p$, write $p(x)$ to mean $p(X = x)$. Similarly, we will write $\mathbf{x}$ to denote the event $\mathbf{X} = \mathbf{x}$, and write $p(\mathbf{x})$ to denote $p(\mathbf{X} = \mathbf{x})$.

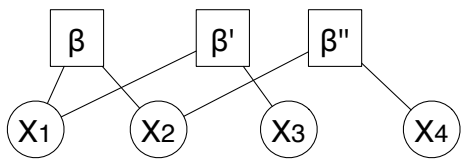

Figure 1: A simple factor graph for $p(\mathbf{x}) = \psi_\beta(x_1, x_2)\psi_{\beta'}(x_1, x_3)\psi_{\beta''}(x_2, x_4)$.

A recurring theme in this paper will be on defining message passing algorithms for joint distributions on factor graphs (Kschischang, Frey, & Loeliger, 2001). In a joint distribution defined as a product of local functions (functions defined on a small subset of variables), we will refer to the local functions as factors. We will index factors, e.g. $\psi_\beta$, with Greek letters, e.g., $\beta, \gamma$ (avoiding $\alpha$ which is used as the symbol for clause to variable ratio in SAT instances). For each factor $\psi_\beta$, we denote $V(\beta) \subseteq V$ as the subset of variables on which $\psi_\beta$ is defined, i.e. $\psi_\beta$ is a function defined on the variables $\mathbf{X}_{V(\beta)}$. In message passing algorithms, messages are vectors of real numbers that are sent from factors to variables or vice versa. A vector message sent from a variable $X_i$ to a factor $\psi_\beta$ will be denoted as $\mathbf{M}_{i \to \beta}$, and a message from $\psi_\beta$ to $X_i$ will be denoted as $\mathbf{M}_{\beta \to i}$.

## 2.2 Joint Distributions and Factor Graphs

Given a large set of discrete, random variables $\mathbf{X} = \{X_i\}_{i \in V}$, we are interested in the joint probability distribution $p(\mathbf{X})$ over these variables. When the set $V$ is large, it is often of interest to assume a simple decomposition, so that we can draw conclusions efficiently from the distribution. In this paper, we are interested in the joint probability distribution that can be decomposed as follows

$$p(\mathbf{X} = \mathbf{x}) = \frac{1}{Z} \prod_{\beta \in F} \psi_\beta(\mathbf{x}_{V(\beta)}), \tag{1}$$

where the set $F$ indexes a set of functions $\{\psi_\beta\}_{\beta \in F}$. Each function $\psi_\beta$ is defined on a subset of variables $\mathbf{X}_{V(\beta)}$ of the set $\mathbf{X}$, and maps configurations $\mathbf{x}_{V(\beta)}$ into non-negative real numbers. Assuming that each function $\psi_\beta$ is defined on a small subset of variables $\mathbf{X}_{V(\beta)}$, we hope to do efficient inference with this distribution, despite the large number of variables in $\mathbf{X}$. The constant $Z$ is a normalization constant, which ensures that the distribution sums to one over all configurations $\mathbf{x}$ of $\mathbf{X}$.

A factor graph (Kschischang et al., 2001) provides a useful graphical representation illustrating the dependencies defined in the joint probability distribution in Equation 1. A factor graph $G = (\{V, F\}, E)$, is a bipartite graph with two sets of nodes, the set of variable nodes, $V$, and the set of factor nodes, $F$. The set of edges $E$ in the factor graph connects variable nodes to factor nodes, hence the bipartite nature of the graph. For a factor graph representing the joint distribution in Equation 1, an edge $e = (\beta, i)$ is in $E$ if and only if the variable $X_i$ is a parameter of the factor $\psi_\beta$ (i.e. $i \in V(\beta)$). We will denote $V(i)$ as the set of factors depending on the variable $X_i$, i.e.

$$V(i) \quad = \quad \{\beta \in F \mid i \in V(\beta)\} \tag{2}$$

We show a simple example of a factor graph in Figure 1. In this small example, we have for example, $V(\beta) = \{1, 2\}$ and $V(2) = \{\beta, \beta''\}$. The factor graph representation is useful for illustrating inference algorithms on joint distributions in the form of Equation 1 (Kschischang et al., 2001). In Section 3, we will describe the BP algorithm by using the factor graph representation.

Equation 1 defines the joint distribution as a product of local factors. It is often useful to represent the distribution in the following exponential form:

$$p(\mathbf{x}) \quad = \quad \exp\Big(\sum_{\beta \in F} \phi_\beta(\mathbf{x}_{V(\beta)}) - \Phi\Big) \tag{3}$$

The above equation is a reparameterization of Equation 1, with $\psi_\beta(\mathbf{x}_{V(\beta)}) = \exp(\phi_\beta(\mathbf{x}_{V(\beta)}))$ and $\Phi = \ln Z$. In statistical physics, the exponential form is often written as follows:

$$p(\mathbf{x}) \quad = \quad \frac{1}{Z} \exp(-\frac{1}{k_B T} E(\mathbf{x})), \tag{4}$$

where $E(\mathbf{x})$ is the Hamiltonian or energy function, $k_B$ is the Boltzmann's constant, and $T$ is the temperature. For simplicity, we set $k_B T = 1$, and Equations 3 and 4 are equivalent with $E(\mathbf{x}) = -\sum_{\beta \in F} \phi_\beta(\mathbf{x}_{V(\beta)})$.

Bayesian (belief) networks and Markov random fields are two other graphical representations often used to describe multi-dimensional probability distributions. Factor graphs are closely related to both Bayesian networks and Markov random fields, and algorithms operating on factor graphs are often directly applicable to Bayesian networks and Markov random fields. We refer the reader to the work of Kschischang et al. (2001) for a comparison between factor graphs, Bayesian networks and Markov random fields.

## 2.3 Inference on Joint Distributions

In the literature, "inference" on a joint distribution can refer to solving one of two problems. We define the two problems as follows:

**Problem 1** (MAP problem). *Given a joint distribution, $p(\mathbf{x})$, we are interested in the configuration(s) with the highest probability. Such configurations, $\mathbf{x}^*$, are called the maximum-a-posteriori configurations, or MAP configurations*

$$\mathbf{x}^* = \arg\max_{\mathbf{x}} p(\mathbf{x}) \tag{5}$$

From the joint distribution in Equation 4, the MAP configuration minimizes the energy function $E(\mathbf{x})$, and hence the MAP problem is sometimes called the **energy minimization** problem.

**Problem 2** (Marginal problem)**.** *Given a joint distribution, $p(\mathbf{x})$, of central interest are the calculation or estimation of probabilities of events involving a single variable $X_i = x_i$. We refer to such probabilities as marginal probabilities:*

$$p_i(x_i) \;\; = \;\; \sum_{\mathbf{x} \backslash x_i} p(\mathbf{x}). \tag{6}$$

The notation $\sum_{\mathbf{x} \backslash x_i}$ means summing over all configurations of $\mathbf{X}$ with the variable $X_i$ set to $x_i$. Marginals are important as they represent the underlying distribution of individual variables.

In general, both problems are not solvable in reasonable time by currently known methods. Naive calculation of $p_i(x_i)$ involves summing the probabilities of all configurations for the variables $\mathbf{X}$ for which $X_i = x_i$. For example, in a factor graph with $n$ variables of cardinality $q$, finding the marginal of one of the variables will involve summing over $q^{n-1}$ configurations. Furthermore, NP-complete problems such as 3-SAT can be simply coded as factor graphs (see Section 2.4.1). As such, the MAP problem is in general NP-complete, while the marginal problem is equivalent to model counting for 3-SAT, and is #P-complete (Cooper, 1990). Hence, in general, we do not expect to solve the inference problems (exactly) in reasonable time, unless the problems are very small, or have special structures that can be exploited for efficient inference.

Of central interest in this paper is a particular approximate inference method known as the (sum-product) belief propagation (BP) algorithm. We defer the discussion of the BP algorithm to the next section. In the rest of this section, we will describe the SAT, Max-SAT and weighted Max-SAT problems, and how they can be simply formulated as inference problems on a joint distribution over a factor graph.

## 2.4 The SAT and Max-SAT Problem

A variable is Boolean if it takes values in {FALSE, TRUE}. In this paper, we will follow conventions in statistical physics, where Boolean variables take values in $\{-1, +1\}$, with $-1$ corresponding to FALSE, and $+1$ corresponding to TRUE.

The Boolean satisfiability (SAT) problem is given as a Boolean propositional formula written with the operators AND (conjunction), OR (disjunction), NOT (negation), and parenthesis. The objective of the SAT problem is to decide whether there exists a configuration such that the propositional formula is satisfied (evaluates to TRUE). The SAT problem is the first problem shown to be NP-complete in Stephen Cook's seminal paper in 1971 (Cook, 1971; Levin, 1973).

The three operators in Boolean algebra are defined as follows: given two propositional formulas $A$ and $B$, OR$(A, B)$ is true if either $A$ or $B$ is true; AND$(A, B)$ is true only if both $A$ and $B$ are true; and NOT$(A)$ is true if $A$ is false. In the rest of the paper, we will use the standard notations in Boolean algebra for the Boolean operators: $A \vee B$ means OR$(A, B)$, $A \wedge B$ means AND$(A, B)$, and $\overline{A}$ means NOT$(A)$. The parenthesis is available to allow nested application of the operators, e.g. $(A \vee B) \wedge (\overline{B} \vee C)$.

The conjunctive normal form (CNF) is often used as a standard form for writing Boolean formulas. The CNF consists of a conjunction of disjunctions of literals, where a literal is either a variable or its negation. For example, $(X_1 \vee \overline{X}_2) \wedge (\overline{X}_3 \vee X_4)$ is in CNF, while $\overline{X_1 \vee X_2}$ and $(X_1 \wedge X_2) \vee (X_2 \wedge X_3)$ are not. Any Boolean formula can be re-written in CNF using De Morgan's law and the distributivity law, although in practice, this may lead to an exponential blowup in the size of the formula, and the Tseitin transformation is often used instead (Tseitin, 1968). In CNF, a Boolean formula can be considered to be the conjunction of a set of clauses, where each clause is a disjunction of literals. Hence, a SAT problem is often given as $(\mathbf{X}, \mathbf{C})$, where $\mathbf{X}$ is the vector of the Boolean variables, and $\mathbf{C}$ is a set of clauses. Each clause in $\mathbf{C}$ is *satisfied* by a configuration if it evaluates to TRUE for that configuration. Otherwise, it is said to be *violated* by the configuration. We will use Greek letters (e.g. $\beta, \gamma$) as indices for clauses in $\mathbf{C}$, and denote by $V(\beta)$ as the set of variables in the clause $\beta \in \mathbf{C}$. The K-SAT problem is a SAT problem for which each clause in $\mathbf{C}$ consists of exactly $K$ literals. The $K$-SAT problem is NP-complete, for $K \geq 3$ (Cook, 1971).

The maximum satisfiability problem (Max-SAT) problem is the optimization version of the SAT problem, where the aim is to minimize the number of violated constraints in the formula. We define a simple working example of the Max-SAT problem that we will use throughout the paper:

**Example 1.** *Define an instance of the Max-SAT problem in CNF with the following clauses* $\beta_1 = (\overline{x}_1 \vee x_2), \beta_2 = (\overline{x}_2 \vee x_3), \beta_3 = (\overline{x}_3 \vee x_1), \beta_4 = (\overline{x}_1 \vee \overline{x}_2 \vee \overline{x}_3), \beta_5 = (x_1 \vee x_2 \vee x_3)$ *and* $\beta_6 = (x_1 \vee x_2)$. *The Boolean expression representing this problem would be*

$$(\overline{x}_1 \vee x_2) \wedge (\overline{x}_2 \vee x_3) \wedge (\overline{x}_3 \vee x_1) \wedge (\overline{x}_1 \vee \overline{x}_2 \vee \overline{x}_3) \wedge (x_1 \vee x_2 \vee x_3) \wedge (x_1 \vee x_2). \tag{7}$$

*The objective of the Max-SAT problem would be to find a configuration minimizing the number of violated clauses.*

### 2.4.1 Factor Graph Representation for SAT Instances

The SAT problem in CNF can easily be represented as a joint distribution over a factor graph. In the following definition, we give a possible definition of a joint distribution over Boolean configurations for a given SAT instance, where the Boolean variables take values in $\{-1, +1\}$.

**Definition 1.** *Given an instance of the SAT problem, $(\mathbf{X}, \mathbf{C})$ in conjunctive normal form, where $\mathbf{X}$ is a vector of $N$ Boolean variables. We define the energy, $E(\mathbf{x})$, and the distribution, $p(\mathbf{x})$, over configurations of the SAT instance (Battaglia et al., 2004)*

$$\forall \beta \in \mathbf{C}, \quad C_\beta(\mathbf{x}_{V(\beta)}) \;=\; \prod_{i \in V(\beta)} \frac{1}{2}(1 + J_{\beta,i} x_i), \tag{8}$$

$$E(\mathbf{x}) \;=\; \sum_{\beta \in C} C_\beta(\mathbf{x}_{V(\beta)}), \tag{9}$$

$$p(\mathbf{x}) \;=\; \frac{1}{Z} \exp(-E(\mathbf{x})), \tag{10}$$

*where $\mathbf{x} \in \{-1, +1\}^N$, and $J_{\beta,i}$ takes values in $\{-1, +1\}$. If $J_{\beta,i} = +1$ (resp. $-1$), then $\beta$ contains $X_i$ as a negative (resp. positive) literal. Each clause $\beta$ is satisfied if one of its variables $X_i$ takes the value $-J_{\beta,i}$. When a clause $\beta$ is satisfied, $C_\beta(\mathbf{x}_{V(\beta)}) = 0$. Otherwise $C_\beta(\mathbf{x}_{V(\beta)}) = 1$.*

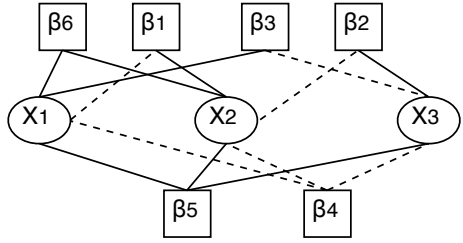

Figure 2: The factor graph for the SAT instance given in Example 1. Dotted (resp. solid) lines joining a variable to a clause means the variable is a negative (resp. positive) literal in the clause.

With the above definition, the energy function is zero for satisfying configurations, and equals the number of violated clauses for non-satisfying configuration. Hence, satisfying configurations of the SAT instance are the MAP configurations of the factor graph.

In this section, we make some definitions that will be useful in the rest of the paper. For a clause $\beta$ containing a variable $X_i$ (associated with the value of $J_{\beta,i}$), we will say that $X_i$ *satisfies* $\beta$ if $X_i = -J_{\beta,i}$. In this case, the clause $\beta$ is satisfied regardless of the values taken by the other variables. Conversely, we say that $X_i$ *violates* $\beta$ if $X_i$ does not satisfy $\beta$. In this case, it is still possible that $\beta$ is satisfied by other variables.

**Definition 2.** *For a clause $\beta \in C$, we define $u_{\beta,i}$ (resp. $s_{\beta,i}$) as the value of $X_i \in \{-1, +1\}$ that violates (resp. satisfies) clause $\beta$. This means that $s_{\beta,i} = -J_{\beta,i}$ and $u_{\beta,i} = +J_{\beta,i}$. We define the following sets*

$$
\begin{aligned}
V^+(i) &= \{\beta \in V(i); s_{\beta,i} = +1\}, \\
V^-(i) &= \{\beta \in V(i); s_{\beta,i} = -1\}, \\
V^s_\beta(i) &= \{\gamma \in V(i) \setminus \{\beta\}; s_{\beta,i} = s_{\gamma,i}\}, \\
V^u_\beta(i) &= \{\gamma \in V(i) \setminus \{\beta\}; s_{\beta,i} \neq s_{\gamma,i}\}.
\end{aligned}
\tag{11}
$$

In the above definitions, $V^+(i)$ (resp. $V^-(i)$) is the set of clauses that contain $X_i$ as a positive literal (resp. negative literal). $V^s_\beta(i)$ (resp. $V^u_\beta(i)$) is the set of clauses containing $X_i$ that agrees (resp. disagrees) with the clause $\beta$ concerning $X_i$. These sets will be useful when we define the SP message passing algorithms for SAT instances.

The factor graph representing the Max-SAT instance given in Example 1 is shown in Figure 2. For this example, $V^+(1) = \{\beta_3, \beta_5, \beta_6\}, V^-(1) = \{\beta_1, \beta_4\}, V^s_{\beta_3}(1) = \{\beta_5, \beta_6\}$, and $V^u_{\beta_3}(1) = \{\beta_1, \beta_4\}$. The energy for this example is as follows:

$$
\begin{aligned}
E(\mathbf{x}) = &\frac{1}{4}(1+x_1)(1-x_2) + \frac{1}{4}(1+x_2)(1-x_3) + \frac{1}{4}(1+x_3)(1-x_1) + \\
&\frac{1}{8}(1+x_1)(1+x_2)(1+x_3) + \frac{1}{8}(1-x_1)(1-x_2)(1-x_3) + \frac{1}{4}(1-x_1)(1-x_2) \quad (12)
\end{aligned}
$$

### 2.4.2 RELATED WORK ON SAT

The SAT problem is well studied in computer science: as the archetypical NP-complete problem, it is common to reformulate other NP-complete problems such as graph coloring as a SAT problem (Prestwich, 2003). SAT solvers are either complete or incomplete. The best known complete solver for solving the SAT problem is probably the Davis-Putnam-Logemann-Loveland (DPLL) algorithm (Davis & Putnam, 1960; Davis, Logemann, & Loveland, 1962). The DPLL algorithm is a basic backtracking algorithm that runs by choosing a literal, assigning a truth value to it, simplifying the formula and then recursively checking if the simplified formula is satisfiable; if this is the case, the original formula is satisfiable; otherwise, the same recursive check is done assuming the opposite truth value. Variants of the DPLL algorithm such as Chaff (Moskewicz & Madigan, 2001), MiniSat (Een & Sörensson, 2005), and RSAT (Pipatsrisawat & Darwiche, 2007) are among the best performers in recent SAT competitions (Berre & Simon, 2003, 2005). Solvers such as *satz* (Li & Anbulagan, 1997) and *cnfs* (Dubois & Dequen, 2001) have also been making progress in solving hard random 3-SAT instances.

Most solvers that participated in recent SAT competitions are complete solvers. While incomplete or stochastic solvers do not show that a SAT instance is unsatisfiable, they are often able to solve larger satisfiable instances than complete solvers. Incomplete solvers usually start with a randomly initialized configuration, and different algorithms differ in the way they flip selected variables to move towards a solution. One disadvantage of such an approach is that in hard SAT instances, a large number of variables have to be flipped to move a current configuration out of a local minimum, which acts as a local trap. Incomplete solvers differ in the strategies used to move the configuration out of such traps. For example, simulated annealing (Kirkpatrick, Jr., & Vecchi, 1983) allows the search to move uphill, controlled by a temperature parameter. GSAT (Selman, Levesque, & Mitchell, 1992) and WalkSAT (Selman, Kautz, & Cohen, 1994) are two algorithms developed in the 1990s that allow randomized moves when the solution cannot be improved locally. The two algorithms differ in the way they choose the variables to flip. GSAT makes the change which minimizes the number of unsatisfied clauses in the new configuration, while WalkSAT selects the variable that, when flipped, results in no previously satisfied clauses becoming unsatisfied. Variants of algorithms such as WalkSAT and GSAT use various strategies, such as tabu-search (McAllester, Selman, & Kautz, 1997) or adapting the noise parameter that is used, to help the search out of a local minima (Hoos, 2002). Another class of approaches is based on applying discrete Lagrangian methods on SAT as a constrained optimization problem (Shang & Wah, 1998). The Lagrange mutlipliers are used as a force to lead the search out of local traps.

The SP algorithm (Braunstein et al., 2005) has been shown to beat the best incomplete solvers in solving hard random 3-SAT instances efficiently. SP estimates marginals on all variables and chooses a few of them to fix to a truth value. The size of the instance is then reduced by removing these variables, and SP is run again on the remaining instance. This iterative process is called decimation in the SP literature. It was shown empirically that SP rarely makes any mistakes in its decimation, and SP solves very large 3-SAT instances that are very hard for local search algorithms. Recently, Braunstein and Zecchina (2006) have shown that by modifying BP and SP updates with a reinforcement term, the effectiveness of these algorithms as solvers can be further improved.

## 2.5 The Weighted Max-SAT Problem

The weighted Max-SAT problem is a generalization of the Max-SAT problem, where each clause is assigned a weight. We define an instance of the weighted Max-SAT problem as follows:

**Definition 3.** *A weighted Max-SAT instance* $(\mathbf{X}, \mathbf{C}, \mathbf{W})$ *in CNF consists of* $\mathbf{X}$, *a vector of* $N$ *variables taking values in* $\{-1, +1\}$, $\mathbf{C}$, *a set of clauses, and* $\mathbf{W}$, *the set of weights for each clause in* $\mathbf{C}$. *We define the energy of the weighted Max-SAT problem as*

$$E(\mathbf{x}) = \sum_{\beta \in C} \prod_{i \in V(\beta)} \frac{w_\beta}{2} (1 + J_{\beta,i} x_i), \tag{13}$$

*where* $\mathbf{x} \in \{-1, +1\}^N$, *and* $J_{\beta,i}$ *takes values in* $\{-1, +1\}$, *and* $w_\beta$ *is the weight of the clause* $\beta$. *The total energy,* $E(\mathbf{x})$, *of a configuration* $\mathbf{x}$ *equals the total weight of violated clauses.*

Similarly to SAT, there are also complete and incomplete solvers for the weighted Max-SAT problem. Complete weighted Max-SAT solvers involve branch and bound techniques by calculating bounds on the cost function. Larrosa and Heras (2005) introduced a framework that integrated the branch and bound techniques into a Max-DPLL algorithm for solving the Max-SAT problem. Incomplete solvers generally employ heuristics that are similar to those used for SAT problems. An example of an incomplete method is the min-conflicts hill-climbing with random walks algorithm (Minton, Philips, Johnston, & Laird, 1992). Many SAT solvers such as WalkSAT can be extended to solve weighted Max-SAT problems, where the weights are used as a criterion in the selection of variables to flip.

As a working example in this paper, we define the following instance of a weighted Max-SAT problem:

**Example 2.** *We define a set of weighted Max-SAT clauses in the following table:*

| Id | Clause | Weight | - - - | - - + | - + - | - + + | + - - | + - + | + + - | + + + |
|---|---|---|---|---|---|---|---|---|---|---|
| $\beta_1$ | $\overline{x}_1 \vee x_2$ | 1 | ✓ | ✓ | ✓ | ✓ | ✗ | ✗ | ✓ | ✓ |
| $\beta_2$ | $\overline{x}_2 \vee x_3$ | 2 | ✓ | ✓ | ✗ | ✓ | ✓ | ✓ | ✗ | ✓ |
| $\beta_3$ | $\overline{x}_3 \vee x_1$ | 3 | ✓ | ✗ | ✓ | ✗ | ✓ | ✓ | ✓ | ✓ |
| $\beta_4$ | $\overline{x}_1 \vee \overline{x}_2 \vee \overline{x}_3$ | 4 | ✓ | ✓ | ✓ | ✓ | ✓ | ✓ | ✓ | ✗ |
| $\beta_5$ | $x_1 \vee x_2 \vee x_3$ | 5 | ✗ | ✓ | ✓ | ✓ | ✓ | ✓ | ✓ | ✓ |
| $\beta_6$ | $x_1 \vee x_2$ | 6 | ✗ | ✗ | ✓ | ✓ | ✓ | ✓ | ✓ | ✓ |
| **Energy** | | | *1* | *9* | *2* | *3* | *1* | *1* | *2* | *4* |

*This weighted Max-SAT example has the same variables and clauses as the Max-SAT example given in Example 1. In the above table, we show the clauses satisfied (a tick) or violated (a cross) by each of the 8 possible configurations of the 3 variables. In the first*

*row, the symbol − corresponds to the value −1, and + corresponds to +1. For example, the string "−−+" corresponds to the configuration $(X_1, X_2, X_3) = (-1, -1, +1)$. The last row of the table shows the energy of the configuration in each column.*

The factor graph for this weighted Max-SAT example is the same as the one for the Max-SAT example in Example 1. The differences between the two examples are in the clause weights, which are reflected in the joint distribution, but not in the factor graph. The energy for this example is as follows:

$$E(\mathbf{x}) = \frac{1}{4}(1 + x_1)(1 - x_2) + \frac{2}{4}(1 + x_2)(1 - x_3) + \frac{3}{4}(1 + x_3)(1 - x_1) +$$

$$\frac{4}{8}(1 + x_1)(1 + x_2)(1 + x_3) + \frac{5}{8}(1 - x_1)(1 - x_2)(1 - x_3) + \frac{6}{4}(1 - x_1)(1 - x_2) \quad (14)$$

## 2.6 Phase Transitions

The SP algorithm has been shown to work well on 3-SAT instances near its phase transition, where instances are known to be very hard to solve. The term "phase transition" arises from the physics community. To understand the notion of "hardness" in optimization problems, computer scientists and physicists have been studying the relationship between computational complexity in computer science and phase transitions in statistical physics. In statistical physics, the phenomenon of phase transitions refers to the abrupt changes in one or more physical properties in thermodynamic or magnetic systems with a small change in the value of a variable such as the temperature. In computer science, it has been observed that in random ensembles of instances such as K-SAT, there is a sharp threshold where randomly generated problems undergo an abrupt change in properties. For example, in K-SAT, it has been observed empirically that as the clause to variable ratio $\alpha$ changes, randomly generated instances change abruptly from satisfiable to unsatisfiable at a particular value of $\alpha$, often denoted as $\alpha_c$. Moreover, instances generated with a value of $\alpha$ close to $\alpha_c$ are found to be extremely hard to solve.

Computer scientists and physicists have worked on bounding and calculating the precise value of $\alpha_c$ where the phase transition for 3-SAT occurs. Using the cavity approach, physicists claim that $\alpha_c \approx 4.267$ (Mezard & Zecchina, 2002). While their derivation of the value of $\alpha_c$ is non-rigorous, it is based on this derivation that they formulated the SP algorithm. Using rigorous mathematical approaches, the upper bounds to the value of $\alpha_c$ can be derived using first-order methods. For example, in the work of Kirousis, Kranakis, Krizanc, and Stamatiou (1998), $\alpha_c$ for 3-SAT was upper bounded by 4.571. Achlioptas, Naor and Peres (2005) lower-bounded the value of $\alpha_c$ using a weighted second moments method, and their lower bound is close to the upper bounds for K-SAT ensembles for large values of $K$. However, their lower bound for 3-SAT is 2.68, rather far from the conjectured value of 4.267. A better (algorithmic) lower bound of 3.52 can be obtained by analyzing the behavior of algorithms that find SAT configurations (Kaporis, Kirousis, & Lalas, 2006).

Physicists have also shown rigorously using second moment methods that as $\alpha$ approaches $\alpha_c$, the search space fractures dramatically, with many small solution clusters appearing relatively far apart from each other (Mezard, Mora, & Zecchina, 2005). Clusters of solutions are generally defined as a set of connected components of the solution space, where two adjacent solutions have a Hamming distance of 1 (differ by one variable). Daude,

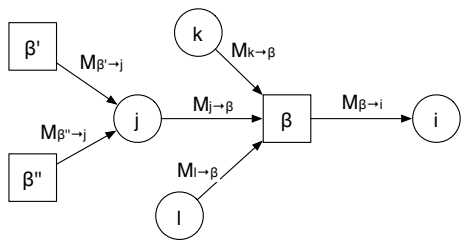

Figure 3: Illustration of messages in a BP algorithm.

Mezard, Mora, and Zecchina (2008) redefined the notion of clusters by using the concept of $x$-satisfiability: a SAT instance is $x$-satisfiable if there exists two solutions differing by $Nx$ variables, where $N$ is the total number of variables. They showed that near the phase transition, $x$ goes from around $\frac{1}{2}$ to very small values, without going through a phase of intermediate values. This clustering phenomenon explains why instances generated with $\alpha$ close to $\alpha_c$ are extremely hard to solve with local search algorithm: it is difficult for the local search algorithm to move from a local minimum to the global minimum.

## 3. The Belief Propagation Algorithm

The BP algorithm has been reinvented in different fields under different names. For example, in the speech recognition community, the BP algorithm is known as the forward-backward procedure (Rabiner & Juang, 1993). On tree-structured factor graphs, the BP algorithm is simply a dynamic programming approach applied to the tree structure, and it can be shown that BP calculates the marginals for each variable in the factor graph (i.e. solving Problem 2). In loopy factor graphs, the BP algorithm has been found to provide a reasonable approximation to solving the marginal problem when the algorithm converges. In this case, the BP algorithm is often called the *loopy* BP algorithm. Yedidia, Freeman and Weiss (2005) have shown that the fixed points of the loopy BP algorithm correspond to the stationary points of the *Bethe* free energy, and is hence a sensible approximate method for estimaing marginals.

In this section, we will first describe the BP algorithm as a dynamic programming method for solving the marginal problem (Problem 2) for tree-structured factor graphs. We will also briefly describe how the BP algorithm can be applied to factor graphs with loops, and refer the reader to the work of Yedidia et al. (2005) for the underlying theoretical justification in this case.

Given a factor graph representing a distribution $p(\mathbf{x})$, the BP algorithm involves iteratively passing messages from factor nodes $\beta \in F$ to variable nodes $i \in V$, and vice versa. Each factor node $\beta$ represents a factor $\psi_\beta$, which is a factor in the joint distribution given in Equation 1. In Figure 3, we give an illustration of how the messages are passed between factor nodes and variable nodes. Each Greek alphabet (e.g. $\beta \in F$) in a square represents a factor (e.g. $\psi_\beta$) and each Roman alphabet (e.g. $i \in V$) in a circle represents a variable (e.g. $X_i$).

The factor to variable messages (e.g. $\mathbf{M}_{\beta \to i}$), and the variable to factor messages (e.g. $\mathbf{M}_{i \to \beta}$) are vectors of real numbers, with length equal to the cardinality of the variable $X_i$.

We denote by $M_{\beta \to i}(x_i)$ or $M_{i \to \beta}(x_i)$ the component of the vector corresponding to the value $X_i = x_i$.

The message update equations are as follows:

$$M_{j \to \beta}(x_j) = \prod_{\beta' \in V(j) \backslash \beta} M_{\beta' \to j}(x_j) \tag{15}$$

$$M_{\beta \to i}(x_i) = \sum_{\mathbf{x}_{V(\beta)} \backslash x_i} \psi_\beta(\mathbf{x}_{V(\beta)}) \prod_{j \in V(\beta) \backslash i} M_{j \to \beta}(x_j), \tag{16}$$

where $\sum_{\mathbf{x}_{V(\beta)} \backslash x_i}$ means summing over all configurations $\mathbf{X}_{V(\beta)}$ with $X_i$ set to $x_i$.

For a tree-structured factor graph, the message updates can be scheduled such that after two parses over the tree structure, the messages will converge. Once the messages converge, the beliefs at each variable node are calculated as follows:

$$B_j(x_j) = \prod_{\beta \in V(j)} M_{\beta \to j}(x_j). \tag{17}$$

For a tree-structured graph, the normalized beliefs for each variable will be equal to its marginals.

---

**INPUT:** A joint distribution $p(\mathbf{x})$ defined over a tree-structured factor graph $(\{V, F\}, E)$, and a variable $X_i \in \mathbf{X}$.

**OUTPUT:** Exact marginals for the variable $X_i$.

**ALGORITHM** :

1. Organize the tree so that $X_i$ is the root of the tree.

2. Start from the leaves, propagate the messages from child nodes to parent nodes right up to the root $X_i$ with Equations 15 and 16.

3. The marginals of $X_i$ can then be obtained as the normalized beliefs in Equation 17.

Figure 4: The BP algorithm for calculating the marginal of a single variable, $X_i$, on a tree-structured factor graph

---

The algorithm for calculating the exact marginals of a given variable $X_i$, is given in Figure 4. This algorithm is simply a dynamic programming procedure for calculating the marginals, $p_i(X_i)$, by organizing the sums so that the sums at the leaves are done first. For the simple example in Figure 1, for calculating $p_1(x_1)$, the sum can be ordered as follows:

$$\begin{aligned} p_1(x_1) &= \sum_{x_2, x_3, x_4} p(\mathbf{x}) \\ &= \psi_\beta(x_1, x_2) \sum_{x_2} . \sum_{x_3} \psi_{\beta'}(x_1, x_3) \sum_{x_4} \psi_{\beta''}(x_2, x_4) \end{aligned}$$

The BP algorithm simply carries out this sum by using the node for $X_1$ as the root of the tree-structured factor graph in Figure 1.

The BP algorithm can also be used for calculating marginals for all variables efficiently, with the message passing schedule given in Figure 5. This schedule involves selecting a random variable node as the root of the tree, and then passing the messages from the leaves to the root, and back down to the leaves, After the two parses, all the message updates required in the algorithm in Figure 4 for any one variable would have been performed, and the beliefs of all the variables can be calculated from the messages. The normalized beliefs for each variable will be equal to the marginals for the variable.

---

**INPUT:** A joint distribution $p(\mathbf{x})$ defined over a tree-structured factor graph $(V, F)$.

**OUTPUT:** Exact marginals for all variables in $V$.

**ALGORITHM** :

1. Randomly select a variable as a root.

2. *Upward pass:* starting from leaves, propagate messages from the leaves right up to the tree.

3. *Downward pass:* from the root, propagate messages back down to the leaves.

4. Calculate the beliefs of all variables as given in Equation 17.

Figure 5: The BP algorithm for calculating the marginals of all variables on a tree-structured factor graph

---

If the factor graph is not tree-structured (i.e. contains loops), then the message updates cannot be scheduled in the simple way described in the algorithm in Figure 5. In this case, we can still apply BP by iteratively updating the messages with Equations 15 and 16, often in a round-robin manner over all factor-variable pairs. This is done until all the messages converge (i.e. the messages do not change over iterations). There is no guarantee that all the messages will converge for general factor graphs. However, if they do converge, it was observed that the beliefs calculated with Equation 17 are often a good approximation of the exact beliefs of the joint distribution (Murphy, Weiss, & Jordan, 1999). When applied in this manner, the BP algorithm is often called the *loopy* BP algorithm. Recently, Yedidia, Freeman and Weiss (2001, 2005) have shown that loopy BP has an underlying variational principle. They showed that the fixed points of the BP algorithm correspond to the stationary points of the Bethe free energy. This fact serves in some sense to justify the BP algorithm even when the factor graph it operates on has loops, because minimizing the Bethe free energy is a sensible approximation procedure for solving the marginal problem. We refer the reader to the work of Yedidia et al. (2005) for more details.

## 4. Survey Propagation: The SP and SP-$y$ Algorithms

Recently, the SP algorithm (Braunstein et al., 2005) has been shown to beat the best incomplete solvers in solving hard 3-SAT instances efficiently. The SP algorithm was first derived from principles in statistical physics, and can be explained using the cavity approach (Mezard & Parisi, 2003). It was first given a BP interpretation in the work of Braunstein and Zecchina (2004). In this section, we will define the SP and the SP-$y$ algorithms for solving SAT and Max-SAT problems, using a warning propagation interpretation for these algorithms.

### 4.1 SP Algorithm for The SAT Problem

In Section 2.4.1, we have defined a joint distribution for the SAT problem $(\mathbf{X}, \mathbf{C})$, where the energy function of a configuration is equal to the number of violated clauses for the configuration. In the factor graph $(\{V, F\}, E)$ representing this joint distribution, the variable nodes in $V$ correspond to the Boolean variables in $\mathbf{X}$, and each factor node in $F$ represents a clause in $\mathbf{C}$. In this section, we provide an intuitive overview of the SP algorithm as it was formulated in the work of Braunstein et al. (2005).

The SP algorithm can be defined as a message passing algorithm on the factor graph $(\{V, F\}, E)$. Each factor $\beta \in F$ passes a single real number, $\eta_{\beta \to i}$ to a neighboring variable $X_i$ in the factor graph. This real number $\eta_{\beta \to i}$ is called a *survey*. According to the warning propagation interpretation given in the work of Braunstein et al. (2005), the survey $\eta_{\beta \to i}$ corresponds to the *probability*[1] of the warning that the factor $\beta$ is sending to the variable $X_i$. Intuitively, if $\eta_{\beta \to i}$ is close to 1, then the factor $\beta$ is warning the variable $X_i$ against taking a value that will violate the clause $\beta$. If $\eta_{\beta \to i}$ is close to 0, then the factor $\beta$ is *indifferent* over the value taken by $X_i$, and this is because the clause $\beta$ is satisfied by other variables in $V(\beta)$.

We first define the messages sent from a variable $X_j$ to a neighboring factor $\beta$, as a function of the inputs from other factors containing $X_j$, i.e. $\{\eta_{\beta' \to j}\}_{\beta' \in V(j) \backslash \beta}$. In SP, this message is a vector of three numbers, $\Pi_{j \to \beta}^u, \Pi_{j \to \beta}^s$, and $\Pi_{j \to \beta}^0$, with the following interpretations:

$\Pi_{j \to \beta}^u$ is the probability that $X_j$ is warned (by other clauses) to take a value that will violate the clause $\beta$.

$\Pi_{j \to \beta}^s$ is the probability that $X_j$ is warned (by other clauses) to take a value that will satisfy the clause $\beta$.

$\Pi_{j \to \beta}^0$ is the probability that $X_j$ is free to take any value.

With these definitions, the update equations are as follows:

$$\Pi_{j \to \beta}^u = \left[1 - \prod_{\beta' \in V_\beta^u(j)} (1 - \eta_{\beta' \to j})\right] \prod_{\beta' \in V_\beta^s(j)} (1 - \eta_{\beta' \to j}), \tag{18}$$

$$\Pi_{j \to \beta}^s = \left[1 - \prod_{\beta' \in V_\beta^s(j)} (1 - \eta_{\beta' \to j})\right] \prod_{\beta' \in V_\beta^u(j)} (1 - \eta_{\beta' \to j}), \tag{19}$$

---

1. SP reasons over clusters of solutions, and the *probability* of a warning in this section is used loosely in the SP literature to refer to the fraction of clusters for which the warning applies. In the next section, we will define a rigorous probability distribution over covers for the RSP algorithm.

$$\Pi^0_{j \to \beta} = \prod_{\beta' \in V(j)} (1 - \eta_{\beta' \to j}), \tag{20}$$

$$\eta_{\beta \to i} = \prod_{j \in V(\beta) - i} \frac{\Pi^u_{j \to \beta}}{\Pi^u_{j \to \beta} + \Pi^s_{j \to \beta} + \Pi^0_{j \to \beta}} \tag{21}$$

These equations are defined using the sets of factors $V^u_\beta(j)$ and $V^s_\beta(j)$, which has been defined in Section 2.4.1. For the event where the variable $X_j$ is warned to take on a value violating $\beta$, it has to be (a) warned by at least one factor $\beta' \in V^u_\beta(j)$ to take on a satisfying value for $\beta'$, and (b) all the other factors in $V^s_\beta(j)$ are not sending warnings. In Equation 18, the probability of this event, $\Pi^u_{j \to \beta}$, is a product of two terms, the first corresponding to event (a) and the second to event (b). The definitions of $\Pi^s_{j \to \beta}$ and $\Pi^0_{j \to \beta}$ are defined in a similar manner. In Equation 21, the final survey $\eta_{\beta \to i}$ is simply the probability of the joint event that all incoming variables $X_j$ are violating the clause $\beta$, forcing the last variable $X_i$ to satisfy $\beta$.

The SP algorithm consists of iteratively running the above update equations until the surveys converge. When the surveys converged, we can then calculate local biases as follows:

$$\Pi^+_j = [1 - \prod_{\beta \in V^+(j)} (1 - \eta_{\beta' \to j})] \prod_{\beta \in V^-(j)} (1 - \eta_{\beta \to j}), \tag{22}$$

$$\Pi^+_j = [1 - \prod_{\beta \in V^-(j)} (1 - \eta_{\beta' \to j})] \prod_{\beta \in V^+(j)} (1 - \eta_{\beta \to j}), \tag{23}$$

$$\Pi^0_j = \prod_{\beta \in V(j)} (1 - \eta_{\beta \to j}), \tag{24}$$

$$W^+_i = \frac{\Pi^+_j}{\Pi^+_j + \Pi^-_j + \Pi^0_j} \tag{25}$$

$$W^-_i = \frac{\Pi^-_j}{\Pi^+_j + \Pi^-_j + \Pi^0_j} \tag{26}$$

To solve an instances of the SAT problem, the SP algorithm is run until it converges, and a few variables that are highly constrained are set to their preferred values. The SAT instance is then reduced to a smaller instance, and SP can be run again on the smaller instance. This continues until SP fails to set any more variables, and in this case, a local search algorithm such as WalkSAT is run on the remaining instance. This algorithm, called the survey inspired decimation algorithm (Braunstein et al., 2005), is given in the algorithm in Figure 6.

## 4.2 The SP-$y$ Algorithm

In contrast to the SP algorithm, the SP-$y$ algorithm's objective is to solve Max-SAT instances, and hence clauses are allowed to be violated, at a price. The SP algorithm can be understood as a special case of the SP-$y$ algorithm, with $y$ taken to infinity (Battaglia et al., 2004). In SP-$y$, a penalty value of $\exp(-2y)$ is multiplied into the distribution for each violated clause. Hence, although the message passing algorithm allows the violation of clauses, but as the value of $y$ increases, the surveys will prefer configurations that violate a minimal number of clauses.

---

**INPUT:** A SAT problem, and a constant $k$.

**OUTPUT:** A satisfying configuration, or report FAILURE.

**ALGORITHM** :

1. Randomly initialize the surveys.

2. Iteratively update the surveys using Equations 18 to 21.

3. If SP does not converge, go to step 7.

4. If SP converges, calculate $W_i^+$ and $W_i^-$ using Equations 25 and 26.

5. Decimation: sort all variables based on the absolute difference $|W_i^+ - W_i^-|$, and set the top $k$ variables to their preferred value. Simplify the instance with these variables removed.

6. If all surveys equal zero, (no variables can be removed in step 5), output the simplified SAT instance. Otherwise, go back to the first step with the smaller instance.

7. Run WalkSAT on the remaining simplified instance, and output a satisfying configuration if WalkSAT succeeds. Otherwise output FAILURE.

Figure 6: The survey inspired decimation (SID) algorithm for solving a SAT problem (Braunstein et al., 2005)

---

The SP-$y$ algorithm can still be understood as a message passing algorithm over factor graphs. As in SP, each factor, $\beta$, passes a survey, $\eta_{\beta \to i}$, to a neighboring variable $X_i$, corresponding to the probability of the warning. To simplify notations, we define $\eta_{\beta \to i}^+$ (resp. $\eta_{\beta \to i}^-$) to be the probability of the warning against taking the value $+1$ (resp. $-1$), and we define $\eta_{\beta \to i}^0 = 1 - \eta_{\beta \to i}^+ - \eta_{\beta \to i}^-$. In practice, since a clause can only warn against either $+1$ or $-1$ but not both, either $\eta_{\beta \to i}^+$ or $\eta_{\beta \to i}^-$ equals zero: $\eta_{\beta \to i}^{J_{\beta,i}} = \eta_{\beta \to i}$, and $\eta_{\beta \to i}^{-J_{\beta,i}} = 0$, where $J_{\beta,i}$ is defined in Definition 1.

Since clauses can be violated, it is insufficient to simply keep track of whether a variable has been warned against a value or not. It is now necessary to keep track of *how many* times the variable has been warned against each value, so that we know *how many* clauses will be violated if the variable was to take a particular value. Let $H_{j \to \beta}^+$ (resp. $H_{j \to \beta}^-$) be the number of times the variable $X_j$ is warned by factors in $\{\beta'\}_{\beta' \in V(j) \backslash \beta}$ against the value $+1$ (resp. $-1$). In SP-$y$, the variable $X_j$ will be forced by $\beta$ to take the value $+1$ if $H_{j \to \beta}^+$ is smaller than $H_{j \to \beta}^-$, and the penalty in this case will be $\exp(-2y H_{j \to \beta}^+)$. In notations used in the work of Battaglia et al. (2004) describing SP-$y$, let $h_{j \to \beta} = H_{j \to \beta}^+ - H_{j \to \beta}^-$.

Battaglia et al. (2004) defined the SP-$y$ message passing equations that calculate the probability distribution over $h_{j \to \beta}$, based on the input surveys,

$$\{\eta_{\beta' \to j}\}_{\beta' \in V(j) \backslash \beta} = \{\eta_{\beta_1 \to j}, \eta_{\beta_2 \to j}, ..., \eta_{\beta_{(|V(j)|-1)} \to j}\}, \tag{27}$$

where $|V(j)|$ refers to the cardinality of the set $V(j)$. The unnormalized distributions $\widetilde{P}_{j\rightarrow\beta}(h)$ are calculated as follows:

$$\widetilde{P}_{j\rightarrow\beta}^{(1)}(h) = \eta_{\beta_1\rightarrow i}^0 \delta(h) + \eta_{\beta_1\rightarrow i}^+ \delta(h-1) + \eta_{\beta_1\rightarrow i}^- \delta(h+1), \quad (28)$$

$$\forall \gamma \in [2, |V(j)|-1], \quad \widetilde{P}_{j\rightarrow\beta}^{(\gamma)}(h) = \eta_{\beta_\gamma\rightarrow i}^0 \widetilde{P}_{j\rightarrow\beta}^{(\gamma-1)}(h)$$
$$+ \eta_{\beta_\gamma\rightarrow i}^+ \widetilde{P}_{j\rightarrow\beta}^{(\gamma-1)}(h-1) \exp\left[-2y\theta(-h)\right]$$
$$+ \eta_{\beta_\gamma\rightarrow i}^- \widetilde{P}_{j\rightarrow\beta}^{(\gamma-1)}(h+1) \exp\left[-2y\theta(h)\right], \quad (29)$$

$$\widetilde{P}_{j\rightarrow\beta}(h) = \widetilde{P}_{j\rightarrow\beta}^{(|V(j)|-1)}(h), \quad (30)$$

where $\delta(h) = 1$ if $h = 0$, and zero otherwise, and $\theta(h) = 1$ if $h \geq 0$, and zero otherwise. The above equations take into account each neighbor of $j$ excluding $\beta$, from $\gamma = 1$ to $\gamma = |V(j)|-1$. The penalties $\exp(-2y)$ are multiplied every time the value of $h_{j\rightarrow\beta}$ decreases in absolute value, as each new neighbor of $X_j$, $\beta_\gamma$, is added. At the end of the procedure, this is equivalent to multiplying the messages with a factor of $\exp(-2y\times\min(H_{j\rightarrow\beta}^+, H_{j\rightarrow\beta}^-))$.

The $\widetilde{P}_{j\rightarrow\beta}(h)$ are then normalized into $P_{j\rightarrow\beta}(h)$ by computing $\widetilde{P}_{j\rightarrow\beta}(h)$ for all possible values of $h$ in $[-|V(j)|+1, |V(j)|-1]$. The message updates for the surveys are as follows:

$$W_{j\rightarrow\beta}^+ = \sum_{h=1}^{|V(j)|-1} P_{j\rightarrow\beta}(h), \quad (31)$$

$$W_{j\rightarrow\beta}^- = \sum_{h=-|V(j)|+1}^{-1} P_{j\rightarrow\beta}(h), \quad (32)$$

$$\eta_{\beta\rightarrow i}^{-J_{\beta,i}} = 0, \quad (33)$$

$$\eta_{\beta\rightarrow i}^{J_{\beta,i}} = \prod_{j\in V(j)\setminus i} W_{j\rightarrow\beta}^{J_{\beta,j}}, \quad (34)$$

$$\eta_{\beta\rightarrow i}^0 = 1 - \eta_{\beta\rightarrow i}^{J_{\beta,i}}, \quad (35)$$

The quantity $W_{j\rightarrow\beta}^+$ (resp. $W_{j\rightarrow\beta}^-$) is the probability of all events warning against the value $+1$ (resp. $-1$). Equation 34 reflects the fact that a warning is sent from $\beta$ to the variable $X_i$ if and only if all other variables in $\beta$ are warning $\beta$ that they are going to violate $\beta$.

When SP-$y$ converges, the preference of each variable is calculated as follows:

$$W_j^+ = \sum_{h=1}^{|V(j)|} P_j(h), \quad (36)$$

$$W_j^- = \sum_{h=-|V(j)|}^{-1} P_j(h), \quad (37)$$

where the $P_j(h)$ are calculated in a similar manner as the $P_{j\rightarrow\beta}(h)$, except that it does not exclude $\beta$ in its calculations.

With the above definitions for message updates, the SP-$y$ algorithm can be used to solve Max-SAT instances by a survey inspired decimation algorithm similar to the one for SP given in the algorithm in Figure 6. At each iteration of the decimation process, the SP-$y$ decimation procedure selects variables to fix to their preferred values based on the quantity

$$b_{fix}(j) = |W_j^+ - W_j^-| \tag{38}$$

In the work of Battaglia et al. (2004), an additional backtracking process was introduced to make the decimation process more robust. This backtracking process allows the decimation procedure to unfix variables already fixed to their values. For a variable $X_j$ fixed to the value $x_j$, the following quantities are calculated:

$$b_{backtrack}(j) = -x_j(W_j^+ - W_j^-) \tag{39}$$

Variables are ranked according to this quantity and the top variables are chosen to be unfixed. In the algorithm in Figure 7, we show the backtracking decimation algorithm for SP-$y$ (Battaglia et al., 2004), where the value of $y$ is either given as input, or can be determined empirically.

---

**INPUT:** A Max-SAT instance and a constant $k$. Optional input: $y_{in}$ and a backtracking probability $r$.

**OUTPUT:** A configuration.

**ALGORITHM** :

1. Randomly initialize the surveys.

2. If $y_{in}$ is given, set $y = y_{in}$. Otherwise, determine the value of $y$ with the bisection method.

3. Run SP-$y$ until convergence. If SP-$y$ converges, for each variable $X_i$, extract a random number $q \in [0, 1]$.

   (a) If $q > r$, sort the variables according to Equation 38 and fix the top $k$ most biased variables.

   (b) If $q < r$ sort the variables according to Equation 39 and unfix the top $k$ most biased variables.

4. Simplify the instance based on step (3). If SP-$y$ converged and return a non-paramagnetic solution (a paramagnetic solution refers to a set of $\{b_{fix}(j)\}_{j \in V}$ that are not biased to any value for all variables), go to step (1).

5. Run weighted WalkSAT on the remaining instance and outputs the best configuration found.

---

Figure 7: The survey inspired decimation (SID) algorithm for solving a Max-SAT instance (Battaglia et al., 2004)

---

## 5. Relaxed Survey Propagation

It was shown (Maneva et al., 2004; Braunstein & Zecchina, 2004) that SP for the SAT problem can be reformulated as a BP algorithm on an extended factor graph. However, their formulation cannot be generalized to explain the SP-$y$ algorithm which is applicable to Max-SAT problems. In a previous paper (Chieu & Lee, 2008), we extended the formulation in the work of Maneva et al. (2004) to address the Max-SAT problem. In this section, we will modify the formulation in our previous paper (Chieu & Lee, 2008) to address the weighted Max-SAT problem, by setting up an extended factor graph on which we run the BP algorithm. In Theorem 3, we show that this formulation generalizes the BP interpretation of SP given in the work of Maneva et al. (2004), and in the main theorem (Theorem 2), we show that running the loopy BP algorithm on this factor graph estimates marginals over covers of configurations violating a set of clauses with minimal total weight.

We will first define the concept of covers in Section 5.1, before defining the extended factor graph in Section 5.2. In the rest of this section, given a weighted Max-SAT problem $(\mathbf{X}, \mathbf{C}, \mathbf{W})$, we will assume that variables in $\mathbf{X}$ take values in $\{-1, +1, *\}$: the third value is a "don't care" state, corresponding to a no-warning message for the SP algorithm defined in the Section 4.

### 5.1 Covers in Weighted Max-SAT

First, we need to define the semantics of the value $*$ as a "don't care" state.

**Definition 4.** *(Maneva et al., 2004) Given a configuration* $\mathbf{x}$*, we say that a variable* $X_i$ *is the unique satisfying variable for a clause* $\beta \in \mathbf{C}$ *if it is assigned* $s_{\beta,i}$ *whereas all other variables* $X_j$ *in the clause are assigned* $u_{\beta,j}$ *(see Definition 2 for the definitions of* $s_{\beta,i}$ *and* $u_{\beta,i}$*). A variable* $X_i$ *is said to be constrained by the clause* $\beta$ *if it is the unique satisfying variable for* $\beta$*. A variable is unconstrained if it is not constrained by any clauses. Define*

$$\mathrm{CON}_{i,\beta}(\mathbf{x}_\beta) = \mathrm{Ind}(x_i \text{ is constrained by } \beta), \tag{40}$$

*where* $\mathrm{Ind}(P)$ *equals 1 if the predicate* $P$ *is true, and 0 otherwise.*

As an illustration, consider the configuration $\mathbf{X} = (+1, -1, -1)$ in Example 2. In this configuration, $X_1 = +1$ is constrained by the clauses $\beta_5$ and $\beta_6$, $X_2 = -1$ is constrained by $\beta_2$, while $X_3 = -1$ is unconstrained: flipping $X_3$ to $+1$ will not violate any additional clauses for the configuration.

In the following definition, we redefine when a configuration taking values in $\{-1, +1, *\}$ satisfies or violates a clauses.

**Definition 5.** *A configuration satisfies a clause* $\beta$ *if and only if (i)* $\beta$ *contains a variable* $X_i$ *set to the value* $s_{\beta,i}$*, or (ii) when at least two variables in* $\beta$ *take the value* $*$*. A configuration violates a clause* $\beta$ *if all the variables* $X_j$ *in* $\beta$ *are set to* $u_{\beta,j}$*. A configuration* $\mathbf{x}$ *is invalid for clause* $\beta$ *if and only if exactly one of the variables in* $\beta$ *is set to* $*$*, and all the other remaining variables in* $\beta$ *are set to* $u_{\beta,i}$*. A configuration is* valid *if it is valid for all clauses in* $\mathbf{C}$*.*

The above definition for invalid configurations reflects the interpretation that the $*$ value is a "don't care" state: clauses containing a variable $X_i = *$ should *already* be satisfied by other variables, and the value of $X_i$ *does not matter.* So $X_i = *$ cannot be the last remaining possibility of satisfying any clause. In the case where a clause contains two variables set to $*$, the clause can be satisfied by either one of these two variables, so the other variable can take the "don't care" value.

We define a partial order on the set of all valid configurations as follows (Maneva et al., 2004):

**Definition 6.** *Let* $\mathbf{x}$ *and* $\mathbf{y}$ *be two valid configurations. We write* $\mathbf{x} \leq \mathbf{y}$ *if* $\forall i$*, (1)* $x_i = y_i$ *or (2)* $x_i = *$ *and* $y_i \neq *$*.*

This partial order defines a lattice, and Maneva et al. (2004) showed that SP is a "peeling" procedure that peels a satisfying configuration to its minimal element in the lattice. A cover is a minimal element in the lattice. In the SAT region, a cover can be defined as follows (Kroc, Sabharwal, & Selman, 2007):

**Definition 7.** *A cover is a valid configuration* $\mathbf{x} \in \{-1, +1, *\}^N$ *that satisfies all clauses, and has no unconstrained variables assigned -1 or +1.*

The SP algorithm was shown to return marginals over covers (Maneva et al., 2004). In principle, there are two kinds of covers: *true* covers which correspond to satisfying configurations, and *false* covers which do not. Kroc et al. (2007) showed empirically that the number of false covers is negligible for SAT instances. For RSP to apply to weighted Max-SAT instances, we introduce the notion of $v$-cover:

**Definition 8.** *A* $v$*-cover is a valid configuration* $\mathbf{x} \in \{-1, +1, *\}^N$ *such that*

1. *the total weight of clauses violated by the configuration equals* $v$*,*

2. $\mathbf{x}$ *has no unconstrained variables assigned -1 or +1.*

Hence the covers defined in Definition 7 are simply $v$-covers with $v = 0$ (i.e. 0-covers).

## 5.2 The Extended Factor Graph

In this section, we will define a joint distribution over an extended factor graph that is positive only over $v$-covers. First, we will need to define functions that will be used to define the factors in the extended factor graph.

**Definition 9.** *For each clause,* $\beta \in \mathbf{C}$*, the following function assigns different values to configurations that satisfy, violate or are invalid (see Definition 5) for* $\beta$*:*

$$\text{VAL}_\beta(\mathbf{x}_{V(\beta)}) = \begin{cases} 1 & \text{if } \mathbf{x}_{V(\beta)} \text{ satisfies } \beta \\ \exp(-w_\beta y) & \text{if } \mathbf{x}_{V(\beta)} \text{ violates } \beta \\ 0 & \text{if } \mathbf{x}_{V(\beta)} \text{ is invalid} \end{cases} \tag{41}$$

In the above definition, we introduced a parameter $y$ in the RSP algorithm, which plays a similar role to the $y$ in the SP-$y$ algorithm. The term $\exp(-w_\beta y)$ is the penalty for violating a clause with weight $w_\beta$.

**Definition 10.** *(Maneva et al., 2004) Given a configuration* $\mathbf{x}$, *we define the parent set* $P_i(\mathbf{x})$ *of a variable* $X_i$ *to be the set of clauses for which* $X_i = x_i$ *is the unique satisfying variable in a configuration* $\mathbf{x}$, *(i.e. the set of clauses constraining* $X_i$ *to its value). Formally,*

$$P_i(\mathbf{x}) = \{\beta \in \mathbf{C} | \, \mathrm{CON}_{i,\beta}(\mathbf{x}_{\mathbf{V}(\beta)}) = 1\} \tag{42}$$

In Example 2, for the configuration $\mathbf{x} = (+1, -1, -1)$, the parent sets are $P_1(\mathbf{x}) = \{\beta_5, \beta_6\}, P_2(\mathbf{x}) = \{\beta_2\}$, and $P_3(\mathbf{x}) = \emptyset$.

Given the weighted Max-SAT instance $(\mathbf{X}, \mathbf{C}, \mathbf{W})$ and its factor graph, $G = (\{V, F\}, E)$, we now construct another distribution with an associated factor graph $G_s = (\{V, F_s\}, E_s)$ as follows. For each $i \in V$, let $P(i)$ be the set of all possible parent sets of the variable $X_i$. Due to the restrictions imposed by our definition, $P_i(\mathbf{x})$ must be contained in either $V^+(i)$ or $V^-(i)$, but not both. Therefore, the cardinality of $P(i)$ is $2^{|V^+(i)|} + 2^{|V^-(i)|} - 1$. Our extended factor graph is defined on set of the variables $\mathbf{\Lambda} = (\Lambda_1, \Lambda_2, ..., \Lambda_n) \in \mathcal{X}_1 \times \mathcal{X}_2 \times ... \times \mathcal{X}_n$, where $\mathcal{X}_i := \{-1, +1, *\} \times P(i)$. Hence this factor graph has the same number of variables as the original SAT instance, but each variable has a large cardinality. Given configurations $\mathbf{x}$ for the SAT instance, we denote configurations of $\mathbf{\Lambda}$ as $\lambda(\mathbf{x}) = \{\lambda_i(\mathbf{x})\}_{i \in V}$, where $\lambda_i(\mathbf{x}) = (x_i, P_i(\mathbf{x}))$.

The definitions given so far define the semantics of valid configurations and parent sets, and in the rest of this section, we will define factors in the extended factor graph $G_s$ to ensure that the above definitions are satisfied by configurations of $\mathbf{\Lambda}$.

The single variable compatibilities ($\Psi_i$) are defined by the following factor on each variable $\lambda_i(\mathbf{x})$:

$$\Psi_i(\lambda_i(\mathbf{x}) = \{x_i, P_i(\mathbf{x})\}) = \begin{cases} 0 & \text{if } P_i(\mathbf{x}) = \emptyset, x_i \neq * \\ 1 & \text{if } P_i(\mathbf{x}) = \emptyset, x_i = * \\ 1 & \text{for any other valid } (x_i, P_i(\mathbf{x})) \end{cases}. \tag{43}$$

The first case in the above definition for $P_i(\mathbf{x}) = \emptyset$ and $x_i \neq *$ corresponds to the case where the variable $X_i$ is unconstrained, and yet takes a value in $\{-1, +1\}$. Valid configurations that are not $v$-covers (with unconstrained variables set to $-1$ or $+1$) have a zero value in the above factor. Hence only $v$-covers have a positive value for these factors. In the last case in the above definition, the validity of $(x_i, P_i(\mathbf{x}))$ simply means that if $x_i = +1$ (resp. $x_i = -1$), $P_i(\mathbf{x}) \subseteq V^+(i)$ (resp. $P_i(\mathbf{x}) \subseteq V^-(i)$.).

The clause compatibilities ($\Psi_\beta$) are:

$$\Psi_\beta(\lambda(\mathbf{x})_{V(\beta)}) = \mathrm{VAL}_\beta(\mathbf{x}_{\mathbf{V}(\beta)}) \prod_{k \in V(\beta)} \mathrm{Ind}\left([\beta \in P_k(\mathbf{x})] = \mathrm{CON}_{\beta,k}(\mathbf{x}_{V(\beta)})\right), \tag{44}$$

where Ind is defined in Definition 4. These clause compatibilities introduce the penalties in $\mathrm{VAL}_\beta(\mathbf{x}_{V(\beta)})$ into the joint distribution. The second term in the above equation enforces that the parent sets $P_k(\mathbf{x})$ are consistent with the definitions of parent sets in Definition 10 for each variable $X_k$ in the clause $\beta$.

The values of $\mathbf{x}$ determines uniquely the values of $\mathbf{P} = \{P_i(\mathbf{x})\}_{i \in V}$, and hence the distribution over $\lambda(\mathbf{x}) = \{x_i, P_i(\mathbf{x})\}_{i \in V}$ is simply a distribution over $\mathbf{x}$.

**Theorem 1.** *Using the notation* $\mathrm{UNSAT}(\mathbf{x})$ *to represent the set of all clauses violated by* $\mathbf{x}$, *the underlying distribution* $p(\mathbf{\Lambda})$ *of the factor graph defined in this section is positive only*

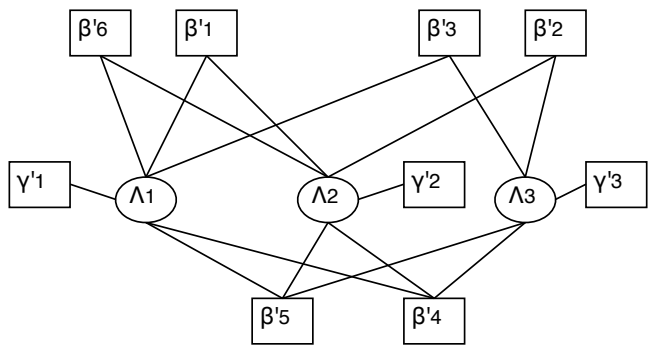

Figure 8: The extended factor graph for the SAT instance given in Example 1. The factor nodes $\beta_i'$ correspond to the clause compatibility factors $\Psi_{\beta_i}$, and the single variable factor nodes $\gamma_i'$ represents the single variable compatibility factors $\Psi_i$. This factor graph is similar to the original factor graph of the SAT instance in Figure 2, except that it has additional factor nodes $\gamma_i'$.

*over v-covers, and for a v-cover* $\mathbf{x}$, *we have:*

$$p(\mathbf{X} = \mathbf{x}) = p(\mathbf{\Lambda} = \lambda(\mathbf{x})) \propto \prod_{\beta \in \text{UNSAT}(\mathbf{x})} \exp(-w_\beta y), \tag{45}$$

*Proof.* Configurations that are not *v*-covers are either invalid or contains unconstrained variables set to $-1$ or $+1$. For invalid configurations, the distribution is zero because of the definition of VAL$_\beta$, and for configurations with unconstrained variables set to $-1$ or $+1$, the distribution is zero due to the definition of the factors $\psi_i$. For each *v*-cover, the total penalty from violated clauses is the product term in Equation 45. $\qquad \square$

The above definition defines a joint distribution over a factor graph. The RSP algorithm is a message passing algorithm defined on this factor graph:

**Definition 11.** *The RSP algorithm is defined as the loopy BP algorithm applied to the extended factor graph* $G_s$ *associated with a MaxSAT instance* $(\mathbf{X}, \mathbf{C}, \mathbf{W})$.

In Section 6, we will formulate the message passing updates for RSP, as well as a decimation algorithm for using RSP as a solver for weighted Max-SAT instances. As an example, Figure 8 shows the extended factor graph for the weighted Max-SAT instance defined in Example 1.

**Definition 12.** *We define a min-cover for a weighted Max-SAT instance as the m-cover, where m is the minimum total weight of violated clauses for the instance.*

**Theorem 2.** *When y is taken to* $\infty$, *RSP estimates marginals over min-covers in the following sense: the stationary points of the RSP algorithm correspond to the stationary points of the Bethe free energy on a distribution that is uniform over min-covers.*

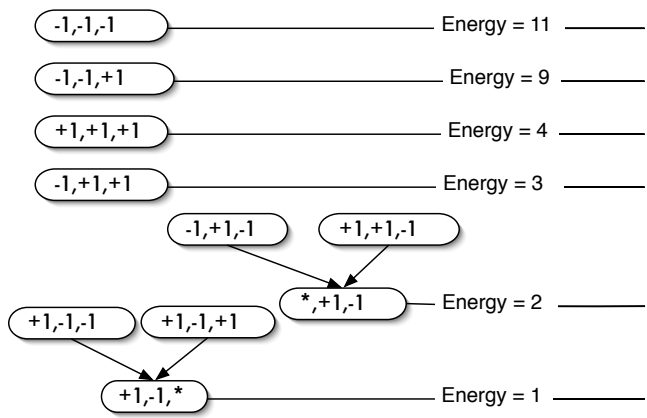

Figure 9: Energy landscape for the weighted Max-SAT instance given in Example 2. Each node represents a configuration for the variables $(x_1, x_2, x_3)$. For example, the node $(-1, +1, -1)$ represents the configuration $(x_1, x_2, x_3) = (-1, +1, -1)$.

*Proof.* The ratio of the probability of a $v$-cover to that of a $(v + \epsilon)$-cover equals $\exp(\epsilon y)$. When $y$ is taken to $\infty$, the distribution in Equation 45 is positive only over min-covers. Hence RSP, as the loopy BP algorithm over the factor graph representing Equation 45, estimates marginals over min-covers. $\square$

In the application of RSP to weighted Max-SAT instances, taking $y$ to $\infty$ would often cause the RSP algorithm to fail to converge. Taking $y$ to a sufficiently large value is often sufficient for RSP to be used to solve weighted Max-SAT instances.

In Figure 9, we show the $v$-covers of a small weighted Max-SAT example in Example 2. In this example, there is a unique min-cover with $X_1 = +1, X_2 = -1$, and $X_3 = *$.

Maneva et al. (2004) formulated the SP-$\rho$ algorithm, which is equivalent to the SP algorithm (Braunstein et al., 2005) for $\rho = 1$. The SP-$\rho$ algorithm is the loopy BP algorithm on the extended factor graph defined in the work of Maneva et al. (2004). Comparing the definitions of the extended factor graph and factors for RSP and SP-$\rho$, we have (Chieu & Lee, 2008):

**Theorem 3.** *By taking $y \to \infty$, RSP is equivalent to SP-$\rho$ with $\rho = 1$.*

*Proof.* The definitions of the joint distribution for SP-$\rho$ for $\rho = 1$ (Maneva et al., 2004), and for RSP in this paper differ only in Definition 9, and with $y \to \infty$ in RSP, their definitions become identical. Since SP-$\rho$ and RSP are both equivalent to the loopy BP on the distribution defined on their extended factor graphs, the equivalence of their joint distribution means that the algorithms are equivalent. $\square$

Taking $y$ to infinity corresponds to disallowing violated clauses, and SP-$\rho$ was formulated for satisfiable SAT instances, where all clauses must be satisfied. For SP-$\rho$, clause weights are inconsequential as all clauses have to be satisfied.

In this paper, we disallow unconstrained variables to take the value $*$. In the Appendix A, we give an alternative definition for the single variable potentials in Equation 43. With this definition, Maneva et al. (2004) defines a smoothing interpretation for SP-$\rho$. This smoothing can also be applied to RSP. See Theorem 6 in the work of Maneva et al. (2004) and the Appendix A for more details.

## 5.3 The Importance of Convergence

It was found that message passing algorithms such as the BP and the SP algorithms perform well whenever they converge (e.g., see Kroc, Sabharwal, & Selman, 2009). While the success of the RSP algorithm on random ensembles of Max-SAT and weighted Max-SAT instances are believed to be due to the clustering phenomenon on such problems, we found that RSP could also be successful in cases where the clustering phenomenon is not observed. We believe that the presence of large clusters help the SP algorithm to converge well, but as long as the SP algorithm converges, the presence of clusters is not necessary for good performance.

When covers are simply Boolean configurations (with no variables taking the $*$ value), they represent singleton clusters. We call such covers *degenerate covers*. In many structured and non random weighted Max-SAT problems, we have found that the covers we found are often degenerate. In a previous paper (Chieu, Lee, & Teh, 2008), we have defined a modified version of RSP for energy minimization over factor graphs, and we show in Lemma 2 in that paper that configurations with * have zero probability, i.e. all covers are degenerate. In that paper, we showed that the value of $y$ can be tuned to favor the convergence of the RSP algorithm.

In Section 7.3, we show the success of RSP on a few benchmark Max-SAT instances. In trying to recover the covers of the configurations found by RSP, we found that all the benchmark instances used have degenerate covers. The fact that RSP converged on these instances is sufficient for RSP to outperform local search algorithms.

## 6. Using RSP for Solving the Weighted Max-SAT Problem

In the previous section, we defined the RSP algorithm in Definition 11 to be the loopy BP algorithm over the extended factor graph. In this section, we will derive the RSP message passing algorithm based on this definition, before giving the decimation-based algorithm used for solving weighted Max-SAT instances.

### 6.1 The Message Passing Algorithm

The variables in the extended factor graphs are no longer Boolean. They are of the form $\lambda_i(\mathbf{x}) = (x_i, P_i(\mathbf{x}))$, which are of large cardinalities. In the definition of the BP algorithm, we have stated that the message vector passed between factors and variables are of length equal to the cardinality of the variables. In this section, we show that the messages passed in RSP can be grouped into a few groups, so that each message passed between variables and factors has only three values.

In RSP, the factor to variable messages are grouped as follows:

$M_{\beta \to i}^s$   if $x_i = s_{\beta,i}, P_i(\mathbf{x}) = S \cup \{\beta\}$, where $S \subseteq V_\beta^s(i)$,

  (all cases where the variable $x_i$ is constrained by the clause $\beta$),

$M^u_{\beta \to i}$  if $x_i = u_{\beta,i}, P_i(\mathbf{x}) \subseteq V^u_\beta(i)$,

(all cases where the variable $x_i$ is constrained to be $u_{\beta,i}$ by other clauses),

$M^{s*}_{\beta \to i}$  if $x_i = s_{\beta,i}, P_i(\mathbf{x}) \subseteq V^s_\beta(i)$,

(all cases where the variable $x_i = s_{\beta,i}$ is not constrained by $\beta$. At least one other variable $x_j$ in $\beta$ satisfies $\beta$ or equals $*$. Otherwise $x_i$ will be constrained),

$M^{**}_{\beta \to i}$  if $x_i = *, P_i(\mathbf{x}) = \emptyset$.

The last two messages are always equal:

$$M^*_{\beta \to i} \quad = \quad M^{s*}_{\beta \to i} = M^{**}_{\beta \to i}.$$

This equality is due to the fact that for a factor that is not constraining its variables, it does not matter whether a variable is satisfying or is $*$, as long as there are at least two variables that are either satisfying or is $*$. In the following, we will consider the two equal messages as a single message, $M^*_{\beta \to i}$.

The variable to factor messages are grouped as follows:

$R^s_{i \to \beta} := \sum_{S \subseteq V^s_\beta(i)} M_{i \to a}(s_{\beta,i}, S \cup \{\beta\})$,

Variable $x_i$ is constrained by $\beta$ to be $s_{\beta,i}$,

$R^u_{i \to \beta} := \sum_{P_i(\mathbf{x}) \subseteq V^u_\beta(i)} M_{i \to a}(u_{\beta,i}, P_i(\mathbf{x}))$,

Variable $x_i$ is constrained by other clauses to be $u_{\beta,i}$,

$R^{s*}_{i \to \beta} := \sum_{P_i(\mathbf{x}) \subseteq V^s_\beta(i)} M_{i \to a}(s_{\beta,i}, P_i(\mathbf{x}))$,

Variable $x_i$ is not constrained by $\beta$, but constrained by other clauses to be $s_{\beta,i}$,

$R^{**}_{i \to \beta} := M_{i \to a}(*, \emptyset)$,

Variable $x_i$ unconstrained and equals *.

The last two messages can again be grouped as one message (as was done in our previous paper, Chieu & Lee, 2008) as follows,

$$R^*_{i \to \beta} \quad = \quad R^{s*}_{i \to \beta} + R^{**}_{i \to \beta},$$

since in calculating the updates of the $M_{\beta \to j}$ messages from the $R_{i \to \beta}$ messages, only $R^*_{i \to \beta}$ is required. The update equations of RSP for weighted Max-SAT are given in Figure 10. These update equations are derived based on loopy BP updates in Equations 15 and 16 in Section 3. In the worst case in a densely connected factor graph, each iteration of updates can be performed in $O(MN)$ time, where $N$ is the number of variables, and $M$ the number of clauses.

### 6.1.1 Factor to Variable Messages

We will begin with the update equations for the messages from factors to variables, given in Equations 46, 47 and 48. The message $M^s_{\beta \to i}$ groups cases where $X_i$ is constrained by

$$M_{\beta \to i}^s = \prod_{j \in V(\beta) \setminus \{i\}} R_{j \to \beta}^u \tag{46}$$

$$M_{\beta \to i}^u = \left[ \prod_{j \in V(\beta) \setminus \{i\}} (R_{j \to \beta}^u + R_{j \to \beta}^*) + \sum_{k \in V(\beta) \setminus \{i\}} (R_{k \to \beta}^s - R_{k \to \beta}^*) \prod_{j \in V(\beta) \setminus \{i,k\}} R_{j \to \beta}^u \right]$$
$$+ (e^{-w_\beta y} - 1) \prod_{j \in V(\beta) \setminus \{i\}} R_{j \to \beta}^u \tag{47}$$

$$M_{\beta \to i}^* = \prod_{j \in V(\beta) \setminus \{i\}} (R_{j \to \beta}^u + R_{j \to \beta}^*) - \prod_{j \in V(\beta) \setminus \{i\}} R_{j \to \beta}^u \tag{48}$$

$$R_{i \to \beta}^s = \prod_{\gamma \in V_\beta^u(i)} M_{\gamma \to i}^u \left[ \prod_{\gamma \in V_\beta^s(i)} (M_{\gamma \to i}^s + M_{\gamma \to i}^*) \right] \tag{49}$$

$$R_{i \to \beta}^u = \prod_{\gamma \in V_\beta^s(i)} M_{\gamma \to i}^u \left[ \prod_{\gamma \in V_\beta^u(i)} (M_{\gamma \to i}^s + M_{\gamma \to i}^*) - \prod_{\gamma \in V_a^u(i)} M_{\gamma \to i}^* \right] \tag{50}$$

$$R_{i \to \beta}^* = \prod_{\gamma \in V_\beta^u(i)} M_{\gamma \to i}^u \left[ \prod_{\gamma \in V_\beta^s(i)} (M_{\gamma \to i}^s + M_{\gamma \to i}^*) - \prod_{\gamma \in V_\beta^s(i)} M_{\gamma \to i}^* \right]$$
$$+ \prod_{\gamma \in V_\beta^s(i) \cup V_\beta^u(i)} M_{\gamma \to i}^* \tag{51}$$

$$B_i(-1) \propto \prod_{\beta \in V^+(i)} M_{\beta \to i}^u \left[ \prod_{\beta \in V^-(i)} (M_{\beta \to i}^s + M_{\beta \to i}^*) - \prod_{\beta \in V^-(i)} M_{\beta \to i}^* \right] \tag{52}$$

$$B_i(+1) \propto \prod_{\beta \in V^-(i)} M_{\beta \to i}^u \left[ \prod_{\beta \in V^+(i)} (M_{\beta \to i}^s + M_{\beta \to i}^*) - \prod_{\beta \in V^+(i)} M_{\beta \to i}^* \right] \tag{53}$$

$$B_i(*) \propto \prod_{\beta \in V(i)} M_{\beta \to i}^* \tag{54}$$

Figure 10: The update equations for RSP. These equations are BP equations for the factor graph defined in the text.

the factor $\beta$. This means that all other variables in $\beta$ are violating the factor $\beta$, and hence we have Equation 46

$$M^s_{\beta \to i} = \prod_{j \in V(\beta) \setminus \{i\}} R^u_{j \to \beta},$$

where $R^u_{j \to \beta}$ are messages from neighbors of $\beta$ stating that they will violate $\beta$.

The next equation for $M^u_{\beta \to i}$ states that the variable $X_i$ is violating $\beta$. In this case, the other variables in $\beta$ are in these possible cases

1. Two or more variables in $\beta$ satisfying $\beta$, with the message update

$$\prod_{j \in V(\beta) \setminus \{i\}} (R^u_{j \to \beta} + R^*_{j \to \beta}) - \sum_{k \in V(\beta) \setminus \{i\}} R^*_{k \to \beta} \prod_{j \in V(\beta) \setminus \{i,k\}} R^u_{j \to \beta} - \prod_{j \in V(\beta) \setminus \{i\}} R^u_{j \to \beta}.$$

2. Exactly one variable in $V(\beta) \setminus \{i\}$ constrained by $\beta$, and all other variables are violating $\beta$, with the message update

$$\sum_{k \in V(\beta) \setminus \{i\}} R^s_{k \to \beta} \prod_{j \in V(\beta) \setminus \{i,k\}} R^u_{j \to \beta}$$

3. All other variables are violating $\beta$, and in this case, there is a penalty factor of $\exp(-w_\beta y)$, with the message update

$$\exp(-w_\beta y) \prod_{j \in V(\beta) \setminus \{i\}} R^u_{j \to \beta}$$

The sum of these three cases result in Equation 48.

The third update equation for $M^*_{\beta \to i}$ is for the case where the variable $X_i$ is unconstrained by $\beta$, satisfying $\beta$ with $s_{\beta,i}$ (for the case $M^{s*}_{\beta \to i}$) or $*$ (for $M^{**}_{\beta \to i}$). This means that there is at least one other satisfying variable that is unconstrained by $\beta$, with the message update

$$\prod_{j \in V(\beta) \setminus \{i\}} (R^u_{j \to \beta} + R^*_{j \to \beta}) - \prod_{j \in V(\beta) \setminus \{i\}} R^u_{j \to \beta}$$

### 6.1.2 Variable to Factor Messages

The first message $R^s_{i \to \beta}$ consists of the case where the variable $X_i$ is constrained by the factor $\beta$, which means that it satisfies neighboring factors in $V^s_\beta(i)$, and violates factors in $V^u_\beta(i)$, with probability

$$\prod_{\gamma \in V^u_\beta(i)} M^u_{\gamma \to i} \left[ \prod_{\gamma \in V^s_\beta(i)} (M^s_{\gamma \to i} + M^{s*}_{\gamma \to i}) \right].$$

The second message $R^u_{i \to \beta}$ is the case where $X_i$ violates $\beta$. In this case, all other variables in $V^u_\beta(i)$ are satisfied, while clauses in $V^s_\beta(i)$ are violated. In this case, the variable $X_i$ must be constrained by one of the clauses in $V_\beta^u(i)$. Hence the message update is

$$\prod_{\gamma \in V_\beta^s(i)} M_{\gamma \to i}^u \left[ \prod_{\gamma \in V_\beta^u(i)} (M_{\gamma \to i}^s + M_{\gamma \to i}^{s*}) - \prod_{\gamma \in V_\beta^u(i)} M_{\gamma \to i}^{s*} \right]$$

The third message $R_{i \to \beta}^*$ is the sum of two messages $R_{i \to \beta}^{s*}$ and $R_{i \to \beta}^{**}$. For the message $R_{i \to \beta}^{s*}$, the variable $X_i$ satisfies $\beta$ but is not constrained by $\beta$, and so it must be constrained by some other factors:

$$\prod_{\gamma \in V_\beta^u(i)} M_{\gamma \to i}^u \left[ \prod_{\gamma \in V_\beta^s(i)} (M_{\gamma \to i}^s + M_{\gamma \to i}^{s*}) - \prod_{\gamma \in V_\beta^s(i)} M_{\gamma \to i}^{s*} \right]$$

The second part of the message, $R_{i \to \beta}^{**}$, is the case where $X_i = *$, :

$$\prod_{\gamma \in V_\beta^s(i) \cup V_\beta^u(i)} M_{\gamma \to i}^{**},$$

and the sum of the above two equations results in Equation 51.

### 6.1.3 THE BELIEFS

The beliefs can be calculated from the factor to variable messages once the algorithm converges, to obtain estimates of the marginals over min-covers. The calculation of the beliefs is similar to the calculation of the variable to factor messages.

The belief $B_i(-1)$ is the belief or the variable $X_i$ taking the value $-1$. This is the case where the variable $X_i$ satisfies clauses in $V^-(i)$, and violates clauses in $V^+(i)$. In this case, $X_i$ must be constrained by one of the factors in $V^-(i)$. Hence the belief is as follows:

$$\prod_{\beta \in V^+(i)} M_{\beta \to i}^u \left[ \prod_{\beta \in V^-(i)} (M_{\beta \to i}^s + M_{\beta \to i}^{s*}) - \prod_{\beta \in V^-(i)} M_{\beta \to i}^{s*} \right].$$

The calculation of the belief $B_i(+1)$ is similar to $B_i(-1)$. The belief $B_i(*)$ is the case where $X_i = *$, and hence it is calculated as follows:

$$\prod_{\beta \in V^(i)} M_{\beta \to i}^{**}.$$

## 6.2 Comparing the RSP and SP-$y$ Message Passing Algorithms

The message passing algorithms for RSP and SP-$y$ share many similarities. Both algorithms

1. include a multiplicative penalty into the distribution for each violated clause.

2. contain a mechanism for a "don't care" state. For SP-$y$, this occurs when a variable receives no warnings from neighboring factors.

However, there are a number of significant differences in the two algorithms.

1. In RSP, the penalties are imposed as each factor passes a message to a variable. For SP-$y$, the penalties are imposed when a variable compiles all the incoming warnings, and decides how many factors it is going to violate.

2. Importantly, in RSP, variables participating in violated clauses can never take the * value. For SP-$y$, a variable receiving an equal number of warnings from the set of factors $\{\beta'\}_{\beta' \in V(i) \setminus \beta}$ against taking the $+1$ and the $-1$ value (i.e. $h_{j \to \beta} = H_{j \to \beta}^{+} - H_{j \to \beta}^{-} = 0$) will decide to pass a message with *no warning* to $\beta$. Hence for SP-$y$, it is possible for variables in violated clauses to take a "don't care" state.

3. In the work of Battaglia et al. (2004) where SP-$y$ was formulated with the cavity approach, it was found that the optimal value of $y$ for a given Max-SAT problem is $y^* = \frac{\delta \Sigma}{\delta e}$, where $\Sigma$ is the complexity in statistical physics, and $e$ is the energy density (Mezard & Zecchina, 2002). They stated that $y^*$ is a finite value when the energy of the Max-SAT problem is not zero. In Theorem 2, we show that for RSP, $y$ should be as large as possible so that the underlying distribution is over min-covers. In our experimental results in Figure 12, we showed that this is indeed true for RSP, as long as it converges.

---

**INPUT:** A (weighted) Max-SAT instance, a constant $k$, and $y_{in}$

**OUTPUT:** A configuration.

**ALGORITHM** :

1. Randomly initialize the surveys and set $y = y_{in}$.

2. Run RSP with $y$. If RSP converges, sort the variables according to the quantities $b_i = |P(x_i = +1) - P(x_i = -1)|$, and fix the top $k$ variables to their preferred values, subject to the condition that $b_i > 0.5$.

3. (For weighted Max-SAT) If RSP fails to converge, adjust the value of $y$.

4. If RSP converges and at least one variable is set, go back to step (1) with the simplified instance. Otherwise, run the (weighted) WalkSAT solver on the simplified instance and output the configuration found.

Figure 11: The decimation algorithm for RSP for solving a (weighted) Max-SAT instance

---

## 6.3 The Decimation Algorithm

The decimation algorithm is shown in Figure 11. This is the algorithm we used for our experiments described in Section 7. In comparing RSP with SP-$y$ on random Max-SAT instances in Section 7.1, we run both algorithms with a fixed $y_{in}$, and vary the $y_{in}$ over a range of values. Comparing Figure 11 to Figure 7 for SP-$y$, the condition used in SP-$y$ to check for a paramagnetic solution is replaced by the condition given in Step (2) in Figure 11. In the experimental results in Section 7.1, we used the SP-$y$ implementation available online (Battaglia et al., 2004), which contains a mechanism for backtracking on decimation decisions (see Figure 7). In Section 7.1, RSP still outperforms SP-$y$ despite not backtracking on its decisions. When running RSP on weighted Max-SAT, we found that it was necessary to adjust $y$ dynamically during the decimation process. For details on experimental settings, please refer to Section 7.

## 7. Experimental Results

We run experiments on random Max-3-SAT, random weighted Max-SAT, as well as on a few benchmark Max-SAT instances used in the work of Lardeux, Saubion, and Hao (2005).

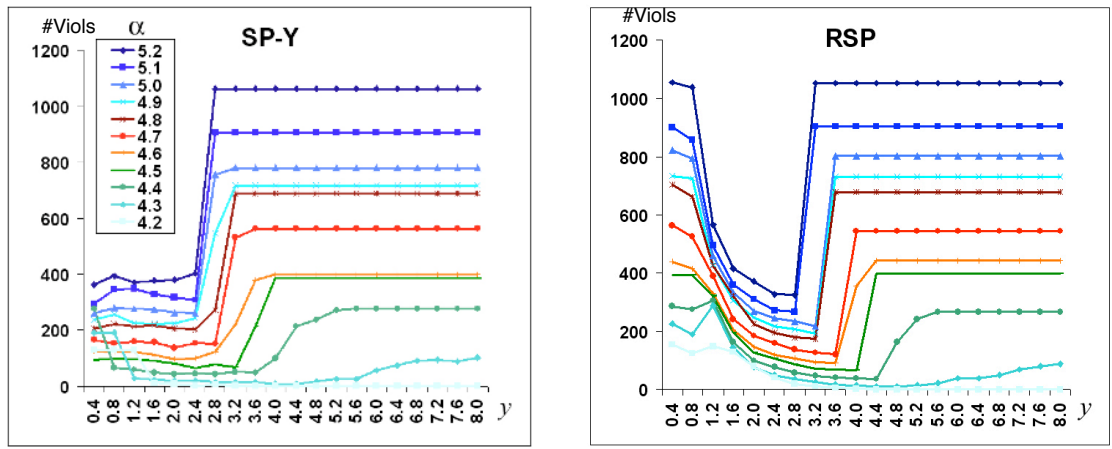

Figure 12: Behaviour of SP-$y$ and RSP over varying values of $y$ on the x-axis, and the number of violated clauses (#viols) on the y-axis. The comparison of the performances between RSP and SP-$y$ are shown in Table 1. The objective of showing the graphs in this figure is to show that the behavior of RSP over varying $y$ is consistent with Theorem 2: as long as RSP converges, its performance improves as $y$ increases. In the graph, RSP reaches a plateau when it fails to converge. This property allows for a systematic search for a good value of $y$ to be used. The behavior of SP-$y$ over varying $y$ is less consistent.

.

### 7.1 Random Max-3-SAT

We run experiments on randomly generated Max-3-SAT instances of $10^4$ variables, with different clause-to-variable ratios. The random instances are generated by the SP-$y$ code available online (Battaglia et al., 2004). In Figure 12, we compare SP-$y$ and RSP on random Max-3-SAT with different clause-to-variable ratio, $\alpha$. We vary $\alpha$ from 4.2 to 5.2 to show the performance of SP-$y$ and RSP in the UNSAT region of 3-SAT, beyond its phase transition at $\alpha_c \approx 4.267$. For each value of $\alpha$, the number of violated clauses (y-axis) is plotted against the value of $y$ used.

We perform the decimation procedure in Figure 11 for RSP, for a fixed value of $y_{in}$, decimating 100 variables at a time (i.e. $k = 100$). For SP-$y$, we run the SP-$y$ code available on line, with the option of decimating 100 variables at each iteration, and with two different settings: with and without backtracking (Battaglia et al., 2004). Backtracking is a procedure used in SP-$y$ to improve performance, by unfixing previously fixed variables at a rate $r = 0.2$, so that errors made by the decimation process can be corrected. For RSP, we do not run backtracking. Note that the $y$ in our formulation equals to $2y$ in the formulation in the work of Battaglia et al. (Battaglia et al., 2004).

Both SP-$y$ and RSP fail to converge when $y$ becomes large enough. When this happens, the output of the algorithm is the result returned by WalkSAT on the original instance. In Figure 12, we see this happening when a curve reaches a horizontal line, signifying that the algorithm is returning the same configuration regardless of $y$ (we "seed" the randomized WalkSAT so that results are identical when instances are identical). From Figure 12, we see RSP performs more consistently than SP-$y$: as $y$ increases, the performance of RSP improves, until a point where RSP fails to converge. Interestingly for Max-3-SAT instances, we observed that once RSP converges for a value of $y$ for a given instance, it will continue to converge for the same value of $y$ throughout the decimation process. Hence, the best value of $y$ for RSP is obtainable without going through the decimation process: we can commence decimation at the largest value of $y$ for which RSP converges. In Table 1, we show that RSP outperforms SP-$y$ for $\alpha \geq 4.7$, despite the fact that we did not allow backtracking for RSP. We also compare RSP and SP-$y$ with the local search solvers implemented in UBCSAT (Tompkins & Hoos, 2004). We run 1000 iterations of each of the 20 Max-SAT solvers in UBCSAT, and take the best result among the 20 solvers. The results are shown in Table 1. We see that the local solvers in UBCSAT does worse than both RSP and SP-$y$. We have also tried running complete solvers such as toolbar (de Givry, Heras, Zytnicki, & Larrosa, 2005) and maxsatz (Li, Manyà, & Planes, 2006). They are unable to deal with instances of size $10^4$.

## 7.2 Random Weighted Max-3-SAT

We have also run experiments on randomly generated weighted Max-3-SAT instances. These instances are generated in the same way as the instances for Max-3-SAT, and in addition, the weights of each clause is uniformly sampled as integers in the set $[1, M]$, where $M$ is the upper bound on the weights. We show the experimental results for $M = 5$ and $M = 10$ in Figure 13. We compare RSP with the 13 weighted Max-SAT solvers implemented in UBCSAT. For RSP, we run all our experiments with an initial $y$ set to 10, and whenever the algorithm fails to converge, we lower the value of $y$ by 1, or halve the value of $y$ if $y$ is less than 1 (see Figure 11). We see that RSP outperforms UBCSAT consistently in all experiments in Figure 13.

## 7.3 Benchmark Max-SAT Instances

We compare RSP with UBCSAT on instances used in the work of Lardeux et al. (2005), which were instances used in the SAT 2003 competition. Among the 27 instances, we use the seven largest instances with more than 7000 variables. We run RSP in two settings: decimating either 10 or 100 variables at a time. We run RSP for increasing values of $y$: for each $y$, RSP fixes a number of spins, and we stop increasing $y$ when the number of spins fixed decreases over the previous value of $y$. For UBCSAT, we run 1000 iterations for each of the 20 solvers. Results are shown in Table 2. Out of the seven instances, RSP fails to fix any spins on the first one, but outperforms UBCSAT on the rest. Lardeux et al. (2005) did not show best performances in their paper, but their average results were an order of magnitude higher than results in Table 2. Figure 12 shows that finding a good $y$ for SP-$y$ is hard. On the benchmark instances, we run SP-$y$ with the "-Y" option (Battaglia et al., 2004) that uses dichotomic search for $y$: SP-$y$ failed to fix any spins on all 7 instances.

Table 1: Number of violated clauses attained by each method. For SP-$y$, "SP-$y$ (BT)" (SP-$y$ with backtracking), and RSP, the best result is selected over all $y$. For each $\alpha$, we show the best performance in bold face. The column "Fix" shows the number of variables fixed by RSP at the optimal $y$, and "Time" the time taken by RSP (in minutes) to fix those variables, on an AMD Opteron 2.2GHz machine.

| $\alpha$ | UBCSAT | SP-$y$ | SP-$y$(BT) | RSP | Fix | Time (minutes) |
|---|---|---|---|---|---|---|
| 4.2 | 47 | **0** | **0** | **0** | 7900 | 24 |
| 4.3 | 68 | 9 | **7** | 10 | 7200 | 43 |
| 4.4 | 95 | 42 | **31** | 36 | 8938 | 82 |
| 4.5 | 128 | 67 | 67 | **65** | 9024 | 76 |
| 4.6 | 140 | 98 | **89** | 90 | 9055 | 45 |
| 4.7 | 185 | 137 | 130 | **122** | 9287 | 76 |
| 4.8 | 232 | 204 | 189 | **172** | 9245 | 52 |
| 4.9 | 251 | 223 | 211 | **193** | 9208 | 62 |
| 5.0 | 278 | 260 | 224 | **218** | 9307 | 66 |
| 5.1 | 311 | 294 | 280 | **267** | 9294 | 42 |
| 5.2 | 358 | 362 | 349 | **325** | 9361 | 48 |

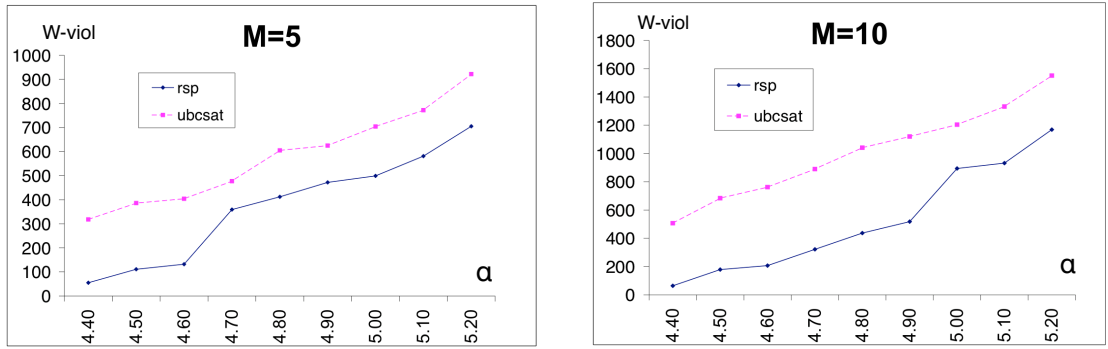

Figure 13: Experimental results for weighted Max-SAT instances. The x-axis shows the value of $\alpha$, and the y-axis (W-viol) is the number of violated clauses returned by each algorithm.

Table 2: Benchmark Max-SAT instances. Columns: "instance" shows the instance name in the paper of Lardeux et al. (2005), "nvar" the number of variables, "ubcsat" and "rsp-$x$" ($x$ is the number of decimations at each iteration) the number of violated clauses returned by each algorithm, and "fx-$x$" the number of spins fixed by RSP. Best results are indicated in bold face.

| instance | nvar | ubcsat | rsp-100 | fx-100 | rsp-10 | fx-10 |
|----------|------|--------|---------|--------|--------|-------|
| family: purdom-10142772393204023 | | | | | | |
| fw | 9366 | **83** | 357 | 0 | 357 | 0 |
| nc | 8372 | 74 | **33** | 8339 | 35 | 8316 |
| nw | 8589 | 73 | **24** | 8562 | 28 | 8552 |
| family: pyhala-braun-unsat | | | | | | |
| 35-4-03 | 7383 | 58 | 68 | 7295 | **44** | 7299 |
| 35-4-04 | 7383 | 62 | 53 | 7302 | **41** | 7304 |
| 40-4-02 | 9638 | 86 | **57** | 9547 | 65 | 9521 |
| 40-4-03 | 9638 | 76 | 77 | 9521 | **41** | 9568 |

The success of the SP family of algorithms on random ensembles of SAT or Max-SAT problem are usually due to the clustering phenomenon on such random ensembles. As the benchmark instances are not random instances, we attempted to see if the configurations found by RSP do indeed belong to a cover representing a cluster of solutions. Rather disappointingly, we found that for all 6 solutions where RSP outperformed local search algorithms, the variables in the solutions are all constrained by at least one clause. Hence, the $v$-covers found are degenerate covers, i.e. the covers do not contain variables set to $*$. It appears that the success of RSP on these benchmark instances is not due to the clustering phenomenon, but simply because RSP manages to converge for these instances, for some value of $y$. Kroc, Sabharwal, and Selman (2009) made a similar observation: the convergence of BP or SP like algorithms is often sufficient for obtaining a good solution to a given problem. As discussed in Section 5.3, the ability to vary $y$ to improve convergence is a useful feature of RSP, but one that is distinct from its ability to exploit the clustering phenomenon.

## 8. Conclusion

While recent work on Max-SAT or weighted Max-SAT tends to focus more on complete solvers, these solvers are unable to handle large instances. In the Max-SAT competition 2007 (Argelich, Li, Manya, & Planes, 2007), the largest Max-3-SAT instances used have only 70 variables. For large instances, complete solvers are still not practical, and local search procedures have been the only feasible alternative. SP-$y$, generalizing SP, has been shown to be able to solve large Max-3-SAT instances at its phase transition, but lacks the theoretical explanations that recent work on SP has generated.

For 3-SAT, there is an easy-hard-easy transition as the clause-to-variable ratio increases. For Max-3-SAT, however, it has been shown empirically that beyond the phase transition of satisfiability, all instances are hard to solve (Zhang, 2001). In this paper, we show that RSP outperforms SP-$y$ as well as other local search algorithms on Max-SAT and weighted Max-SAT instances, well beyond the phase transition region.

Both RSP and SP-$y$ do well on Max-SAT instances near the phase transition. The mechanisms behind SP-$y$ and RSP are similar: both algorithms impose a penalty term for each violated constraint, and both reduce to SP when $y \to \infty$. SP-$y$ uses a population dynamics algorithm, which can also be seen as a warning propagation algorithm. In this paper, we have formulated the RSP algorithm as a BP algorithm over an extended factor graph, enabling us to understand RSP as estimating marginals over min-covers.

## Acknowledgments

This work is supported in part by NUS ARF grant R-252-000-240-112.

## Appendix A. Smoothing Interpretation for RSP

In the definition of SP-$\rho$ (Maneva et al., 2004), the parameter $\rho$ was introduced to define a whole family of algorithms. For $\rho = 1$, the SP-$\rho$ algorithm corresponds to the SP algorithm, while for $\rho = 0$, the SP-$\rho$ algorithm corresponds to the BP algorithm. In this section, we develop a more general version of the extended factor graph defined in Section 5, that incorporates the $\rho$ in SP-$\rho$. We will call the corresponding RSP algorithm on this new factor graph the RSP-$\rho$ algorithm.

The only difference between the factor graph for RSP-$\rho$ and the one in Section 5 is the definition of the variable compatibilities in Equation 43. Following notations in the work of Maneva et al. (2004), we introduce the parameters $\omega_0$ and $\omega_*$, and we restrict ourselves to the case where $\omega_0 + \omega_* = 1$ (The $\rho$ in SP-$\rho$ or RSP-$\rho$ is equal to $\omega_*$). We redefine the variable compatibilities as follows

$$\Psi_i(\lambda_i(\mathbf{x}) = \{x_i, P_i(\mathbf{x})\}) = \begin{cases} \omega_0 & \text{if } P_i(\mathbf{x}) = \emptyset, x_i \neq * \\ \omega_* & \text{if } P_i(\mathbf{x}) = \emptyset, x_i = * \\ 1 & \text{for any other valid } (x_i, P_i(\mathbf{x})) \end{cases}, \tag{55}$$

with $\omega_0 + \omega_* = 1$. The definition in Equation 43 corresponds to the particular case where $\omega_0 = 0$ and $\omega_* = 1$. In Section 5, we have defined the factor graph so that unconstrained variables must take the value $*$. With the new definition of $\Psi_i$ above, unconstrained variables are allowed to take on the values $-1$ or $+1$ with weight $\omega_0$, and the $*$ value with weight $\omega_*$.

With the above definition, the joint distribution in Equation 45 is redefined as follows:

$$P(\mathbf{x}) = P(\{x_k, P_k\}_k) \propto \omega_0^{n_0(\mathbf{x})} \omega_*^{n_*(\mathbf{x})} \prod_{\beta \in \text{UNSAT}(\mathbf{x})} \exp(-w_\beta y). \tag{56}$$

where $n_0(\mathbf{x})$ is the number of unconstrained variables in $\mathbf{x}$ taking $+1$ or $-1$, and $n_*(\mathbf{x})$ the number of unconstrained variables taking $*$ in $\mathbf{x}$.

**Case** $\omega_* = 1$: we have studied this case in the main paper: the underlying distribution is a distribution which is positive only over $v$-covers.

**Case** $\omega_* = 0$**:** in this case, only configurations $\mathbf{x}$ with $n_*(\mathbf{x}) = 0$ have non-zero probability in the distribution given in Equation 56. Hence, the value $*$ is forbidden, and all variables take values in $-1, +1$. A Boolean configuration violating clauses with total weight $W$ has a probability proportional to $\exp(-yW)$. Hence we retreive the weighted Max-SAT energy defined in Equation 13. In this case, the factor graph is equivalent to the original weighted Max-SAT factor graph defined in Definition 3, and hence RSP-$\rho$ is equivalent to the loopy BP algorithm on the original weighted Max-SAT problem.

**Case** $\omega_* \neq 1$ **and** $\omega_* \neq 0$**:** in this case, all valid configurations $\mathbf{x}$ violating clauses with a total weight $W$ has a probability proportional to $\omega_0^{n_0(\mathbf{x})} \omega_*^{n_*(\mathbf{x})} \exp(-yW)$. Hence, the probability of $v$-covers in the case where $\omega_* = 1$ are spread over the lattice for which it is the minimal element.

With the above formulation, RSP-$\rho$ can be seen as a family of algorithms that include the BP and the RSP algorithm, moving from BP to RSP as $\rho$ (or $\omega_*$) varies from 0 to 1.